\documentclass{article}
\usepackage{sketchssm,times}

\usepackage{amsmath,amsfonts,bm}

\def\eqref#1{equation~\ref{#1}}

\def\1{\bm{1}}

\DeclareMathAlphabet{\mathsfit}{\encodingdefault}{\sfdefault}{m}{sl}
\SetMathAlphabet{\mathsfit}{bold}{\encodingdefault}{\sfdefault}{bx}{n}

\DeclareMathOperator*{\argmin}{arg\,min}

\usepackage{booktabs}   %
\usepackage{amsmath}    %
\usepackage{multirow}
\usepackage{pifont}

\usepackage[utf8]{inputenc} %
\usepackage[T1]{fontenc}    %
\usepackage{hyperref}       %
\usepackage{url}            %
\usepackage{booktabs}       %
\usepackage{amsfonts}       %
\usepackage{nicefrac}       %
\usepackage{microtype}      %
\usepackage{xcolor}         %

\newcommand{\name}{{SketchSSM}\xspace}

\usepackage{xspace}
\usepackage{listings}
\usepackage{placeins}
\usepackage{graphicx}

\newcommand{\para}[1]{\noindent \textbf{#1.}}

\usepackage{amssymb}              %

\usepackage{amsmath}              %

\usepackage{mathtools}            %

\usepackage{multirow}
\usepackage{threeparttable}  %

\usepackage{algorithm}
\usepackage{algpseudocode}
\usepackage{enumitem}

\DeclareRobustCommand{\state}[1][0]{\ensuremath{S_{#1}}}

\DeclareRobustCommand{\sketch}{\ensuremath{U}}
\DeclareRobustCommand{\sketchingmatrix}{\ensuremath{\Omega}}

\DeclareRobustCommand{\coefficientvector}[1][t]{\ensuremath{c_{#1}}}
\DeclareRobustCommand{\coefficientmap}{\ensuremath{C}}
\DeclareRobustCommand{\effectivequery}[1][t]{\ensuremath{\tilde q_{#1}}}

\DeclareRobustCommand{\stateoutput}[1][t]{\ensuremath{o_{#1}^{\mathrm{state}}}}
\DeclareRobustCommand{\approxstateoutput}[1][t]{\ensuremath{\hat{o}_{#1}^{\mathrm{state}}}}
\DeclareRobustCommand{\bufferoutput}[1][t]{\ensuremath{o_{#1}^{\mathrm{buffer}}}}

\makeatletter
\newcommand{\DeclarePaperTerm}[2]{%
  \expandafter\newcommand\csname paper@term@#1\endcsname{#2}}
\DeclarePaperTerm{state}{state}
\DeclarePaperTerm{sketch}{sketch}
\DeclarePaperTerm{sketchingmatrix}{sketching matrix}
\DeclarePaperTerm{coefficientvector}{coefficient vector}
\DeclarePaperTerm{coefficientmap}{coefficient map}
\DeclarePaperTerm{effectivequery}{effective query}
\DeclarePaperTerm{stateoutput}{state-read output}
\DeclarePaperTerm{approxstateoutput}{approximate state-read output}
\DeclarePaperTerm{bufferoutput}{buffer output}
\DeclarePaperTerm{reconstructionerror}{reconstruction error}
\DeclarePaperTerm{foldedsketch}{transformed sketch}

\newcommand{\termname}[1]{\csname paper@term@#1\endcsname}
\DeclareRobustCommand{\named}[1]{\termname{#1}~\csname #1\endcsname{}}
\makeatother

\AtBeginDocument{}

\usepackage{hyperref}
\usepackage{url}
\usepackage{adjustbox}
\usepackage{makecell}
\usepackage{pifont}

\title{SketchSSM: Write to the Full State, Read from a Compact Sketch}

\author{%
\begin{tabular}{@{}c@{}}
\textbf{Omin Kwon}\textsuperscript{1}\enspace
\textbf{JoongWon Shin}\textsuperscript{1}\enspace
\textbf{Minseo Kim}\textsuperscript{2}\enspace
\textbf{Kurt Keutzer}\textsuperscript{2}\enspace
\textbf{Sehoon Kim}\textsuperscript{3,\scalebox{1.2}{\ensuremath{\dagger}}}\enspace
\textbf{Jae W. Lee}\textsuperscript{1,\scalebox{1.2}{\ensuremath{\dagger}}}\\[6pt]
\textsuperscript{1}Seoul National University\quad
\textsuperscript{2}UC Berkeley\quad
\textsuperscript{3}KAIST\\[4pt]
{\small\texttt{om0127@snu.ac.kr}\quad
\texttt{leodal@snu.ac.kr}\quad
\texttt{minseo.kim@berkeley.edu}}\\
{\small\texttt{keutzer@berkeley.edu}\quad
\texttt{sehoonkim@kaist.ac.kr}\quad
\texttt{jaewlee@snu.ac.kr}}
\end{tabular}%
}

\hypersetup{
  pdftitle={SketchSSM: Write to the Full State, Read from a Compact Sketch},
  pdfauthor={Omin Kwon, JoongWon Shin, Minseo Kim, Kurt Keutzer, Sehoon Kim, Jae W. Lee}
}

\begin{document}

\maketitle
\begingroup
\renewcommand{\thefootnote}{%
  \scalebox{1.2}{\ensuremath{\dagger}}%
}
\footnotetext[2]{Co-corresponding authors: Sehoon Kim and Jae W. Lee.}
\endgroup

\begin{abstract}
Hybrid-attention models replace most softmax attention layers
with linear attention, reducing KV-cache growth and enabling
larger decode batches where recurrent-state access becomes
a major bottleneck.
ReplaySSM amortizes state updates by buffering keys and values,
but each new query still requires a full-state read even though
the state remains unchanged between state updates.
We observe that low-rank state-weighted query approximation accurately
preserves state-read outputs.
Although future queries are unknown, the basis vectors
used to approximate them can be fixed offline.
Based on this observation, we introduce \name{},
which preserves full-state updates while approximating reads.
At each state update, \name{} reads the full state once
to precompute outputs for these basis vectors,
storing them in a compact sketch.
Each subsequent decode step combines the sketch vectors
with query-dependent coefficients to reconstruct
the output without a full-state read.
Across four Mamba-2-, GDN-, and KDA-based models,
\name{} at a mean sketch rank of 8 reduces state-access traffic by approximately
10$\times$ while matching the average accuracy of the FP32 full-state
baseline across four decode benchmarks, and preserves recall
on four RULER retrieval tasks.
At this rank on one NVIDIA B300, linear-attention kernel speedups
over the Standard vLLM baseline reach
7.30$\times$, 5.02$\times$, and 5.24$\times$
for Mamba-2, GDN, and KDA, respectively,
with up to 2.77$\times$ higher decode throughput
on Nemotron 3 Super.
\par\nobreak\medskip
\begingroup
\small
\hypersetup{hidelinks}
\setlength{\tabcolsep}{4pt}
\renewcommand{\arraystretch}{1.2}
\newcommand{\resourcelinkrow}[4]{%
  \href{#3}{\textbf{#1}} &
  \href{#3}{\raisebox{-0.35ex}{#2}} &
  \href{#3}{\nolinkurl{#4}}}
\noindent\makebox[\linewidth][c]{%
\begin{tabular}{@{}rcl@{}}
\resourcelinkrow{Project Page}{\includegraphics[height=2.2ex]{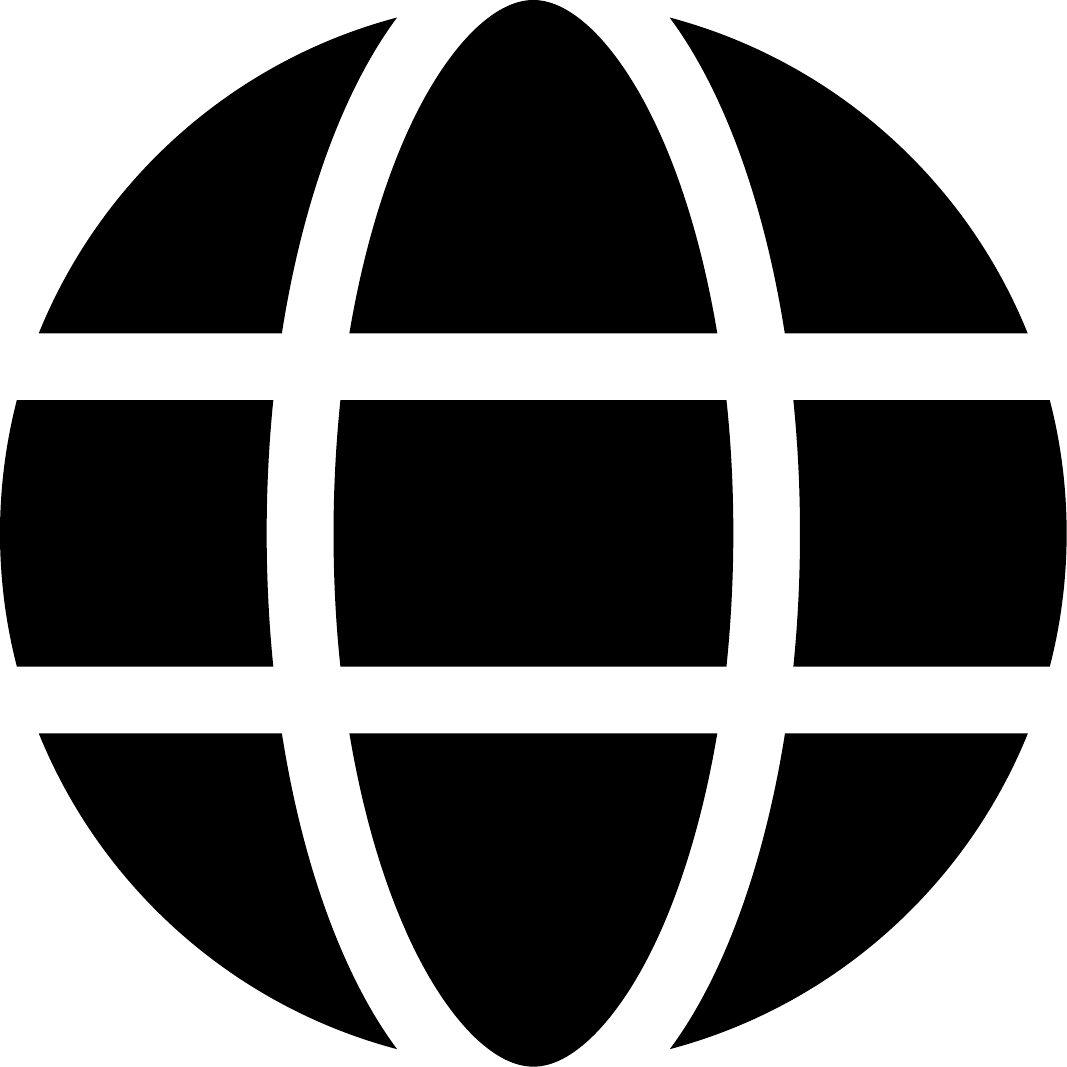}}{https://omin-kwon.github.io/project/SketchSSM}{omin-kwon.github.io/project/SketchSSM}\\
\resourcelinkrow{Code}{\includegraphics[height=2.2ex]{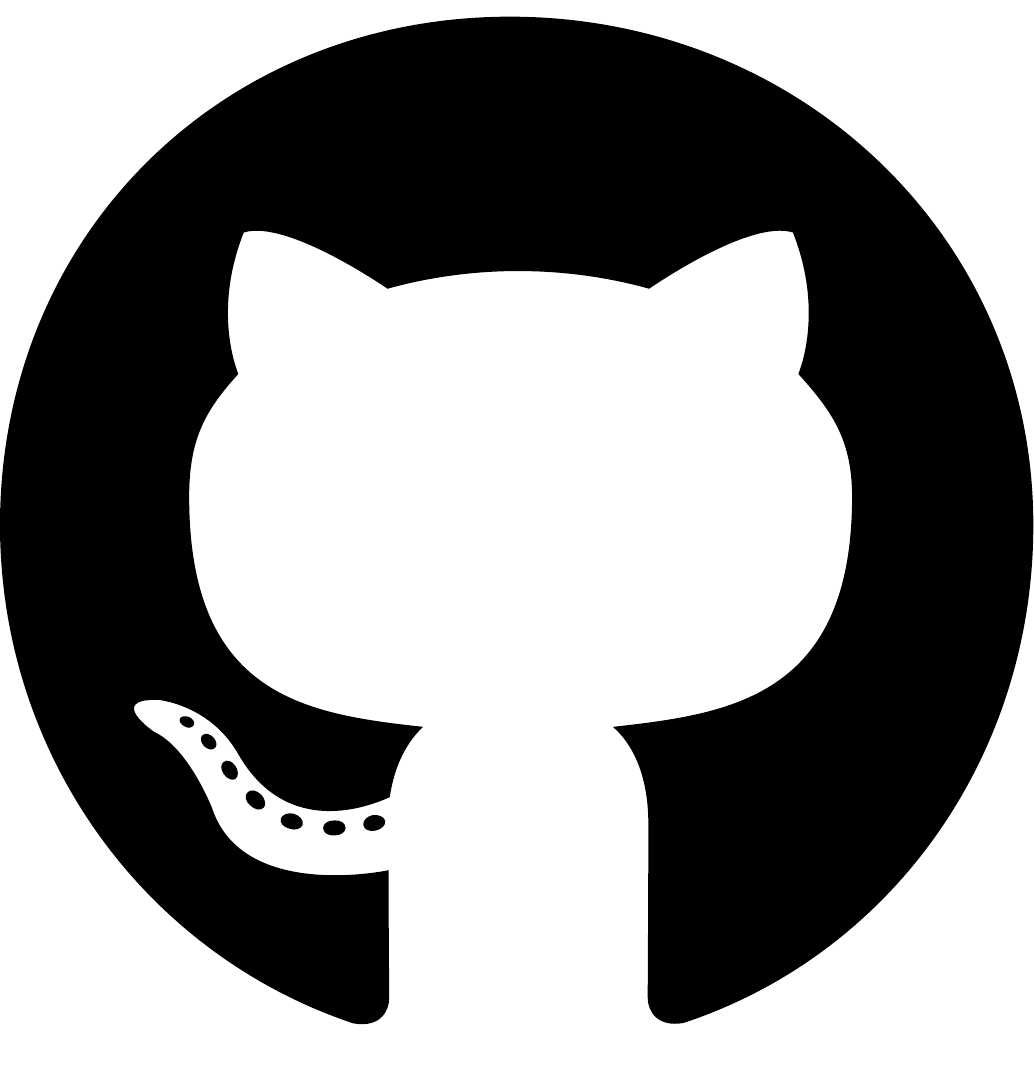}}{https://github.com/SNU-ARC/SketchSSM}{github.com/SNU-ARC/SketchSSM}\\
\resourcelinkrow{Calibration Data}{\includegraphics[height=2.2ex]{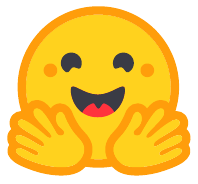}}{https://huggingface.co/SketchSSM}{huggingface.co/SketchSSM}\\
\resourcelinkrow{PyPI}{\includegraphics[height=2.2ex]{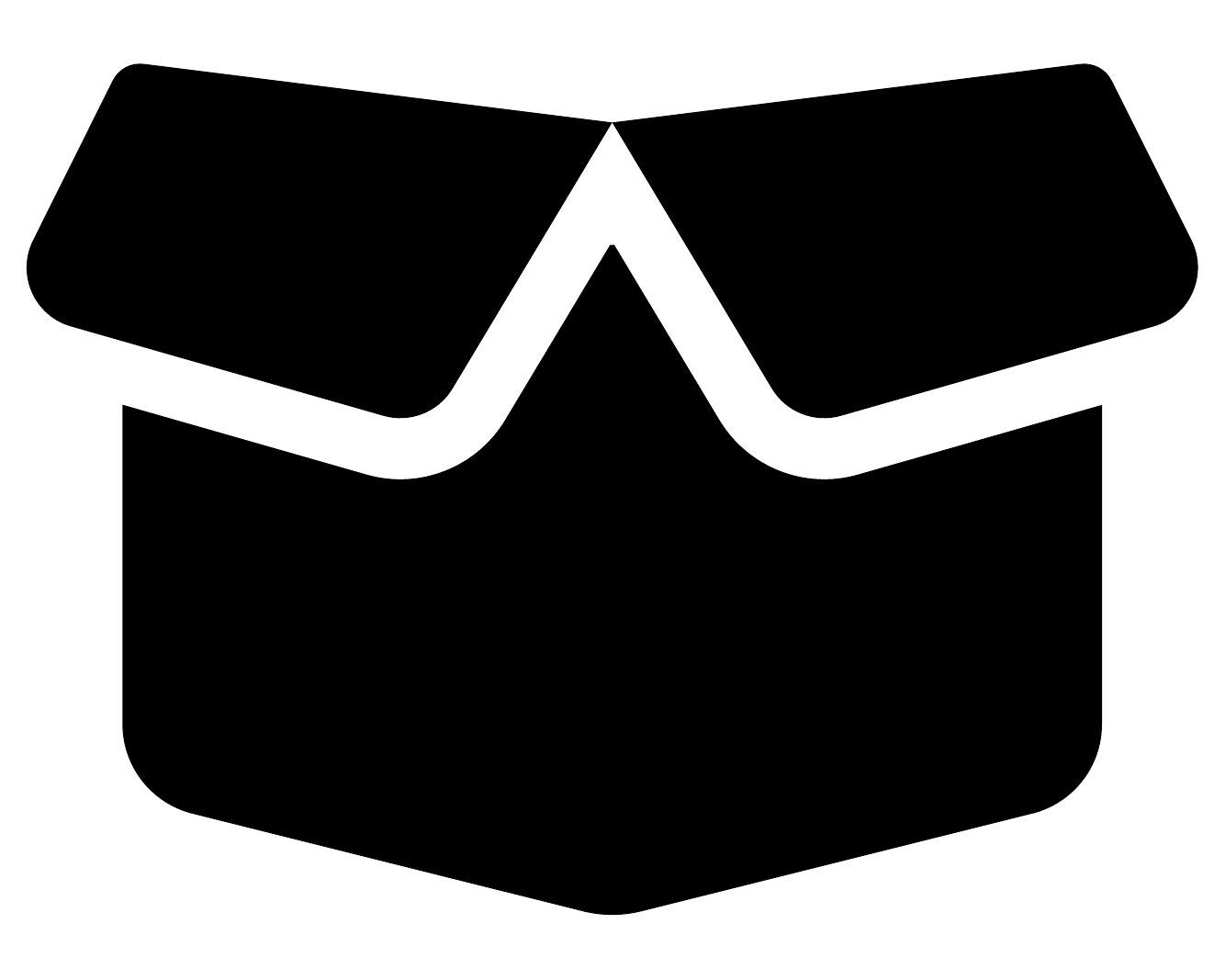}}{https://pypi.org/project/sketchssm/}{pypi.org/project/sketchssm/}
\end{tabular}%
}\par
\endgroup

\end{abstract}

\section{Introduction}

\begin{figure}[t]
\centering
\includegraphics[width=\textwidth]{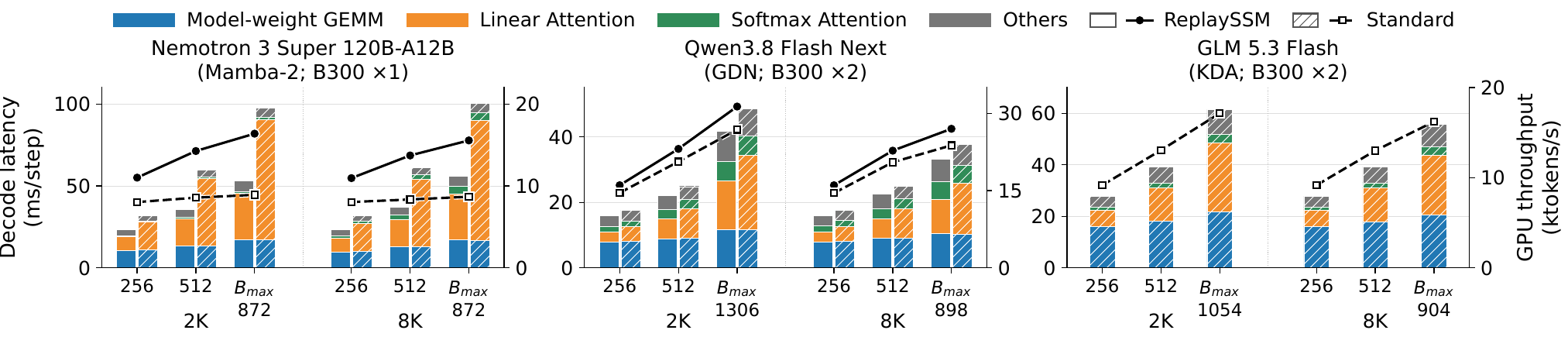}
\caption{Decode latency breakdown and decode throughput
at batch sizes $256$, $512$, and $B_{\max}$ for $2$K and $8$K contexts.
At $B_{\max}$, linear attention accounts for $42$--$75\%$
of decode latency with Standard, and $31$--$53\%$
with ReplaySSM.
$B_{\max}$ denotes the largest tested batch size that fits
within GPU memory for each model and context length. Throughput generally increases as batch size grows toward $B_{\max}$.
GLM uses only Standard because KDA lacks an official
ReplaySSM implementation.}
\label{fig:decode-breakdown}
\end{figure}

Recent hybrid-attention models replace most softmax attention layers with linear attention while retaining high accuracy~\citep{nvidia2026nemotron3super,qwen3.8flashnext,glm5team2026glm5vibecodingagentic}. Replacing KV caches with fixed-size recurrent states reduces cache memory, enabling larger decode batches with higher throughput. However, in this large-batch regime, reading and updating the entire recurrent state every decode step becomes a major decode bottleneck as shown in Fig.~\ref{fig:decode-breakdown}. A common approach to reducing this traffic is to reduce the recurrent state size. Quantization lowers state precision~\citep{q-mamba,chiang2025quamba2,zhang2026dampdecayawaremixedprecisionrecurrentstate}, while pruning removes state entries~\citep{menezes2026ghost,nazari2026keystatereductionlinear}. However, reducing the state size introduces state approximation errors that propagate and accumulate through subsequent decode steps, resulting in substantial accuracy degradation, as we show in Section~\ref{sec:acc_evaluation}.

To reduce \textit{state-write} traffic associated with state updates, ReplaySSM~\citep{daolab2026replayssm} and KVBuffer~\citep{zou2026kvbufferioawareservinglinear} buffer the much smaller keys and values and apply their accumulated updates to the full state once every several decode steps, thereby amortizing state writes. However, \textit{state-read} traffic remains unchanged: computing each output vector requires multiplying the new query by the full state. Therefore, a full-state read occurs at every decode step even though the state remains fixed until the buffered updates are applied. As shown in Fig.~\ref{fig:decode-breakdown}, linear attention remains a major bottleneck even after applying ReplaySSM due to this repeated state-read overhead.

We identify a key opportunity to avoid repeated full-state
reads while accurately reconstructing state-read outputs.
Although the query changes at every decode step,
the state-read output can be approximated using fixed query basis vectors with reconstruction adapted to the current state and query.
We observe that even a small number of query basis vectors
is sufficient to reconstruct outputs with low error,
as shown in Fig.~\ref{fig:sketch-subspace}.

SketchSSM exploits this opportunity by using a single full-state
read at each update step to multiply the state by all fixed query
basis vectors and precompute their outputs.
We store these output vectors in a compact matrix called a \emph{sketch}.
At each decode step between updates, SketchSSM approximates
the new query's output as a linear combination of sketch vectors
with query-dependent coefficients, without reading the full state.
With $G$ fixed query basis vectors and a $K$-dimensional query,
the sketch is only $G/K$ the size of the full state,
where $G\ll K$ (Equation~\ref{equantion-sketch}).
Reading this compact sketch instead of the full state substantially
reduces state-read traffic, as shown in Fig.~\ref{fig:punchline_plot}.
By approximating only reads while preserving exact full-state updates,
SketchSSM avoids propagating state-compression errors through updates.

Across four Mamba-2-, GDN-, and KDA-based models,
SketchSSM at a mean sketch rank of $\bar G=8$ reduces state-access traffic by approximately
10$\times$ while matching the average accuracy of the FP32 full-state baseline
across four benchmarks.
At this rank in vLLM on one NVIDIA B300 GPU, SketchSSM achieves linear-attention kernel speedups of up to 7.30$\times$, 5.02$\times$, and 5.24$\times$ over the Standard full-state baseline for Mamba-2, GDN, and KDA, respectively (3.24$\times$, 3.43$\times$, and 3.69$\times$ over ReplaySSM), and up to 2.77$\times$ higher decode throughput on Nemotron 3 Super 120B-A12B (Section~\ref{eval:speedup}).
Our primary contributions can be summarized as follows:
\begin{itemize}[leftmargin=*,nosep]
    \item We propose a training-free design that preserves
    the full state for updates while approximating
    only state-read outputs, reducing read traffic without
    propagating state-compression errors (Section~\ref{sec:mot-obs}).\vspace{2mm}
    \item We design a sketch of precomputed output vectors
    that can be constructed with a single full-state read.
    We improve accuracy by allocating sketch ranks differently across
    state heads (Section~\ref{sec:sketch}).\vspace{2mm}
    \item We demonstrate substantially better accuracy--traffic tradeoffs than state pruning and quantization, and accelerate linear attention across Mamba-2, GDN, and KDA with optimized kernels (Section~\ref{sec:evaluation}).
\end{itemize}

\section{Related Work}

\para{Recurrent Linear Attention and Hybrid Models}
Linear attention computes recurrent updates and reads over
a fixed-size matrix state, keeping storage independent of
context length~\citep{katharopoulos2020transformers,pmlr-v235-yang24ab}.
Mamba-2 connects structured SSMs and linear attention through state-space duality, while GDN and KDA extend this recurrence with delta-rule erasure and more expressive gating~\citep{dao2024transformers,yang2025gated,kimiteam2025kimi,schlag2021linear}.
Recent hybrid-attention models interleave these recurrent layers with fewer softmax or sparse-attention layers, limiting context-growing KV caches while retaining token-level retrieval~\citep{lieber2024jamba,nvidia2026nemotron3super, qwen2026design,glm5team2026glm5vibecodingagentic}.

\para{Efficient Recurrent-State Inference}
\label{sec:bg-state-access}
The performance bottleneck of linear attention is state access: reading and updating the state at every decode step. State quantization~\citep{chiang2025quamba2,q-mamba,zhang2026dampdecayawaremixedprecisionrecurrentstate} and pruning~\citep{menezes2026ghost,nazari2026keystatereductionlinear} reduce traffic by shrinking the state, but limit information capacity and propagate compression errors through updates~\citep{arora2024simple,qin2024hgrn2,pan2025scalinglinearattentionsparse}. Recently, ReplaySSM~\citep{daolab2026replayssm} and KVBuffer~\citep{zou2026kvbufferioawareservinglinear} reduce write traffic by buffering keys and values and applying accumulated updates at each flush, leaving read traffic unchanged.
SketchSSM complements buffered updates by reducing the remaining read traffic with a sketch.

\para{Matrix Sketching}
Matrix sketching constructs compact representations for approximate
matrix computations~\citep{halko2011finding,woodruff2014sketching},
including randomized projections, streaming sketches, and learned
sketches~\citep{liberty2013simple,indyk2019learning}.
In LLM inference, low-rank methods use activation statistics
to approximate weights~\citep{yuan2023asvd,wang2025svdllm},
or perform low-rank attention to reduce KV-cache
storage~\citep{saxena2024eigenattention}.
SketchSSM builds on the range-finding framework
of \citet{halko2011finding}, as detailed in Appendix~\ref{app:sketch-problem}.
It sketches the state with an offline-calibrated
query basis and reconstructs outputs with
coefficients minimizing output error for the current state and query.

\section{Motivation}
\label{sec:motivation}

\subsection{State-Access Bottleneck in Hybrid-Attention Decoding}
\label{sec:mot-breakdown}

Recent hybrid-attention models substantially reduce decode latency
while retaining high accuracy~\citep{nvidia2026nemotron3super,qwen3.8flashnext,glm5team2026glm5vibecodingagentic}.
Most layers replace context-growing KV caches with fixed-size
recurrent states, allowing larger decode batches.
Fig.~\ref{fig:decode-breakdown} profiles this large-batch regime
at $2$K- and $8$K-token inputs on Nemotron 3 Super (Mamba-2, one B300),
Qwen3.8 Flash-Next (GDN, two B300s), and GLM 5.3 Flash (KDA, two B300s),
all using NVFP4 weights.
It separates each decode step into model-weight GEMMs,
linear attention, softmax attention, and other costs.
Softmax attention takes only a small fraction of decode latency
because these models replace most softmax attention layers with
linear attention and optimize the remaining ones.
Nemotron 3 Super uses an 8-bit KV cache, Qwen3.8 Flash-Next uses
Qwen Sparse Attention (QSA), and GLM 5.3 Flash combines Multi-head
Latent Attention (MLA) with DeepSeek Sparse Attention (DSA);
QSA and DSA bound per-step KV reads during long-context serving.

Linear attention emerges as the dominant decode bottleneck,
as each layer reads and updates an independent recurrent state
for every request at each decode step.
With Standard execution, which updates the full recurrent
state at every decode step, linear attention accounts for
$42$--$75\%$ of decode latency at $B_{\max}$ across the three models.
ReplaySSM buffers keys and values over a window of decode steps
to amortize state-write traffic and lower linear-attention latency,
but leaves state-read traffic unchanged.
Further reducing this bottleneck therefore requires reducing
state-read traffic.

\subsection{Why Every Decode Step Reads the Full State}
\label{sec:mot-decomp}

\para{Reformulation: Decomposing the State Read}
To examine this state-read traffic, consider a $W$-step
buffering window starting from state
$\state\in\mathbb{R}^{K\times V}$,
with $K$ key channels and $V$ value channels.
Let $D_t$ denote the decay matrix and
$D_{a:b}:=D_b\cdots D_a$ its accumulated product.
The state recurrence and output decomposition used by
ReplaySSM~\citep{daolab2026replayssm} and
KVBuffer~\citep{zou2026kvbufferioawareservinglinear}
can be written as
{\small
\begin{align}
\state[t]
&= D_t\state[t-1]+k_tw_t^\top
 = D_{1:t}\state
 + \sum_{s=1}^{t}
   \big(D_{s+1:t}k_s\big)w_s^\top
&& \text{(update)} \label{eq:update} \\
o_t
&= \state[t]^\top q_t
 = {o'}_t^{\mathrm{state}}+{o'}_t^{\mathrm{buffer}},
\quad
{o'}_t^{\mathrm{state}} := \state^\top D_{1:t}^\top q_t,
\quad
{o'}_t^{\mathrm{buffer}} := \sum_{s=1}^{t}
\big\langle D_{s+1:t}k_s,q_t\big\rangle w_s
&& \text{(read)} \label{eq:read}
\end{align}
}
Here, $q_t,k_t\in\mathbb{R}^{K}$ are the query and key,
$v_t,w_t\in\mathbb{R}^{V}$ are the value and write vectors,
and $\beta_t$ is the write gate.
The state term ${o'}_t^{\mathrm{state}}$ is computed directly
from $\state$.
While ${o'}_t^{\mathrm{buffer}}$ is computed from the buffer,
its write vectors
$w_t=\beta_t(v_t-\state[t-1]^\top D_t^\top k_t)$
require reading $\state$ in delta-rule models.
To isolate the effect of reading $\state$,
we collect all state dependence into a single output term.
To this end, we absorb the delta-rule erase into the transition:
$M_t=(I-\beta_t k_tk_t^\top)D_t$ for GDN and KDA,
and $M_t=D_t$ for Mamba-2.
We define $M_{a:b}:=M_b\cdots M_a$ and the effective query
$\effectivequery:=M_{1:t}^\top q_t$.
An adapted WY representation~\citep{yang2024parallelizing}
evaluates $\effectivequery$ through parallel accumulation over
buffered tokens (Appendix~\ref{app:wy-effective-query}).
The update and read can then be written as
{\small
\begin{align}
\state[t]
&= M_t\state[t-1]+\beta_t k_t v_t^\top
 = M_{1:t}\state
 + \sum_{s=1}^{t}\beta_s
   \big(M_{s+1:t}k_s\big)v_s^\top
&& \text{(update)} \label{eq:transition-update} \\
o_t
&= \state[t]^\top q_t
 = \stateoutput+\bufferoutput,
\quad
\stateoutput := \state^\top\effectivequery,
\quad
\bufferoutput := \sum_{s=1}^{t}\beta_s
\big\langle M_{s+1:t}k_s,q_t\big\rangle v_s
&& \text{(read)} \label{eq:m-reads}
\end{align}
}
All dependence on $\state$ is now confined to $\stateoutput$,
while $\bufferoutput$ is computed exactly from raw buffered inputs.
Thus, computing $\stateoutput=\state^\top\effectivequery$
is the source of full-state read traffic at each decode step.

\para{Source of State Read: Changing Queries over the Same State}
Although $\state$ remains fixed throughout the window,
the effective query $\effectivequery=M_{1:t}^\top q_t$
changes at each decode step as both the query $q_t$
and the accumulated transition $M_{1:t}$ change.
Directly evaluating $\stateoutput=\state^\top\effectivequery$
thus requires reading all $K\times V$ state entries
for each query.
The challenge is to answer these changing queries
without repeatedly reading the same full state.

\subsection{Replacing Full-State Reads with a Compact Sketch}
\label{sec:mot-obs}

\para{Observation: A Small, Fixed Query Basis for Accurate State Reads}
We observe that a small, fixed query basis can accurately
reconstruct state-read outputs for changing queries,
using state- and query-dependent coefficients.
Specifically, an effective query $\effectivequery\in\mathbb R^K$
can be approximated as $\sketchingmatrix\coefficientvector$ in a $G$-dimensional subspace,
where $\sketchingmatrix\in\mathbb R^{K\times G}$ contains
$G$ query basis vectors and $\coefficientvector\in\mathbb R^G$
gives the coefficients.
Here, the goal is to preserve the state-read output $\stateoutput$,
rather than the effective query itself.
We therefore measure the approximation quality through the
output reconstruction $\approxstateoutput=\state^\top\sketchingmatrix\coefficientvector$
and define its error as
\begin{equation}
\mathcal E_t^{\mathrm{state}}
:=
\left\|\stateoutput-\approxstateoutput\right\|_2^2
=
\left\|
\state^\top\effectivequery
-
\state^\top\sketchingmatrix\coefficientvector
\right\|_2^2,
\qquad
\state^\top\effectivequery\in\mathbb R^V.
\label{eq:m-output-error}
\end{equation}
Expanding the squared norm, we can rewrite the output error
as a query approximation error weighted by the state Gram
matrix $\state\state^\top$:
{\small
\begin{equation}
\mathcal E_t^{\mathrm{state}}
=
(\effectivequery-\sketchingmatrix\coefficientvector)^\top
\state\state^\top
(\effectivequery-\sketchingmatrix\coefficientvector)
=
\left\|
(\state\state^\top)^{1/2}\effectivequery
-
(\state\state^\top)^{1/2}\sketchingmatrix\coefficientvector
\right\|_2^2,
\qquad
(\state\state^\top)^{1/2}\effectivequery\in\mathbb R^K.
\label{eq:state_weighted_query}
\end{equation}
}

\begin{figure}[t]
\centering
\includegraphics[width=\textwidth]{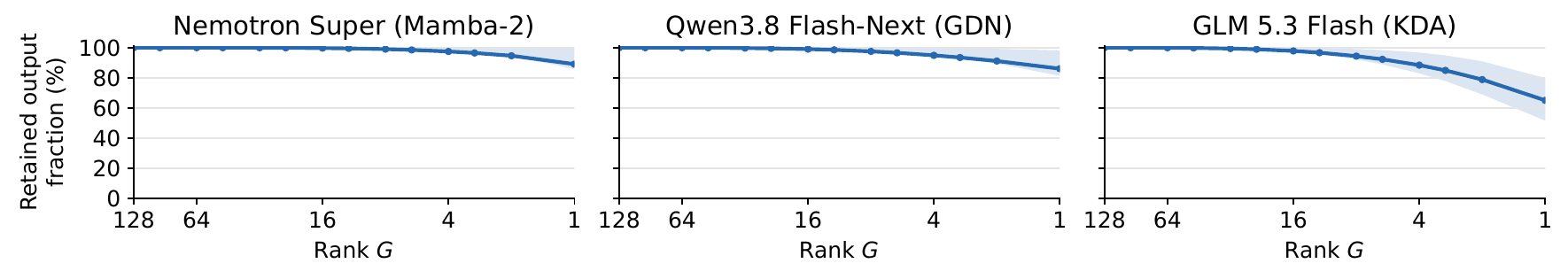}
\caption{Retained state-read output fraction $\rho_G$ versus
rank $G$ on Nemotron 3 Super (Mamba-2), Qwen3.8 Flash-Next (GDN),
and GLM 5.3 Flash (KDA), using a query basis fixed offline
and state-dependent coefficients.
Lines show the mean across heads; shaded regions denote the interquartile range.
At $G=4$, the mean retained output fraction is 88.4--97.5\%
across the three models.
}
\label{fig:sketch-subspace}
\end{figure}

Thus, minimizing output reconstruction error is equivalent to minimizing the state-weighted query approximation error.
This means that larger query errors can be tolerated along directions that the state amplifies less.
Our key observation is that the state-weighted query
$(\state\state^\top)^{1/2}\effectivequery$
can be well approximated within the low-dimensional subspace
spanned by $(\state\state^\top)^{1/2}\sketchingmatrix$,
using state- and query-dependent coefficients
$\coefficientvector$.
Section~\ref{sec:sketch} details how we select the fixed query basis
$\sketchingmatrix$ and compute coefficients
$\coefficientvector$ to minimize output reconstruction error.
Fig.~\ref{fig:sketch-subspace} supports this observation
using the retained output fraction
$\rho_G:=1-\mathbb E_{\mathcal D}[\mathcal E_t^{\mathrm{state}}]/
\mathbb E_{\mathcal D}[\|\stateoutput\|_2^2]$,
computed per head and summarized by the mean across heads.
Using just $G=4$ basis vectors in the $K=128$-dimensional
query space retains 88.4--97.5\% of output energy on
held-out samples, averaged across heads for each of
the three model families.

\para{Key Idea: Precomputing Output Vectors as a Sketch}
This observation allows us to compute $\stateoutput$ accurately while avoiding repeated reads of the full
state for each new query.
With the query approximation
$\effectivequery\approx\sketchingmatrix\coefficientvector$,
we can express the output as
\begin{equation}
\stateoutput
=
\state^\top\effectivequery
\approx
\state^\top\sketchingmatrix\coefficientvector
=
(\state^\top\sketchingmatrix)\coefficientvector
=
\sketch\coefficientvector,
\qquad
\sketch:=\state^\top\sketchingmatrix\in\mathbb R^{V\times G}.
\label{equantion-sketch}
\end{equation}

\begin{figure}[t]
\centering
\includegraphics[width=\textwidth]{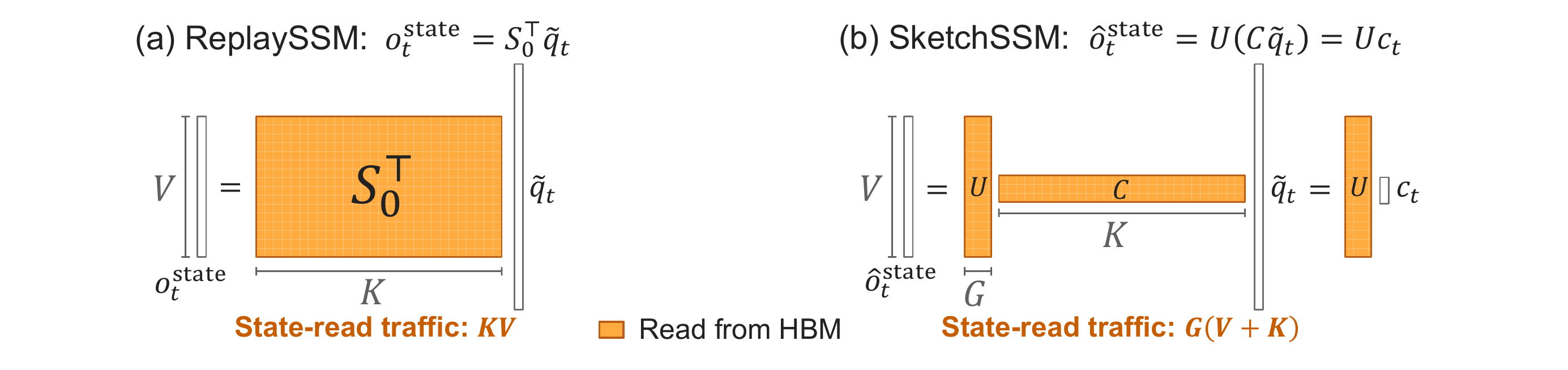}
\caption{
(a) ReplaySSM reads the full state to compute
\(\stateoutput\), incurring \(KV\) elements of state-read traffic.
(b) SketchSSM reconstructs \(\approxstateoutput\) from rank-\(G\) factors,
reducing traffic to \(G(K+V)\) elements.}
\label{fig:punchline_plot}
\end{figure}

The approximated state-read output is then
$\approxstateoutput=\sketch\coefficientvector$.
With fixed $\sketchingmatrix$, a single read of the full
state $\state$ at the flush step suffices to precompute the outputs
for all $G$ query basis vectors in $\Omega$ before the actual queries arrive.
Each column $u_g=\state^\top\omega_g\in\mathbb R^V$
stores the output for one query basis vector.
We call this matrix of $G$ precomputed output vectors
the \emph{sketch}.
Using the sketch $\sketch$, each decode step between state updates
no longer needs to read the full state.
Instead, it reconstructs the output as a linear combination
of output vectors in the sketch using the query-dependent
coefficients $\coefficientvector=\coefficientmap\effectivequery$,
where $\coefficientmap\in\mathbb R^{G\times K}$ is a
coefficient map precomputed from the current state \(S_0\)
per window.
Thus, as illustrated in Fig.~\ref{fig:punchline_plot},
repeated reads of the full state $\state$ are replaced
by reads of the much smaller sketch $\sketch$ and
coefficient map $\coefficientmap$, reducing state-read traffic.
This design decouples read approximation from updates:
the sketch serves only reads, while the full state is updated
exactly from the buffered inputs.
Section~\ref{sec:sketch} next describes how we compute
$\sketchingmatrix$, $\sketch$, and $\coefficientvector$.

\section{SketchSSM}
\label{sec:sketch}

\begin{figure}[t]
\centering
\includegraphics[width=\textwidth]{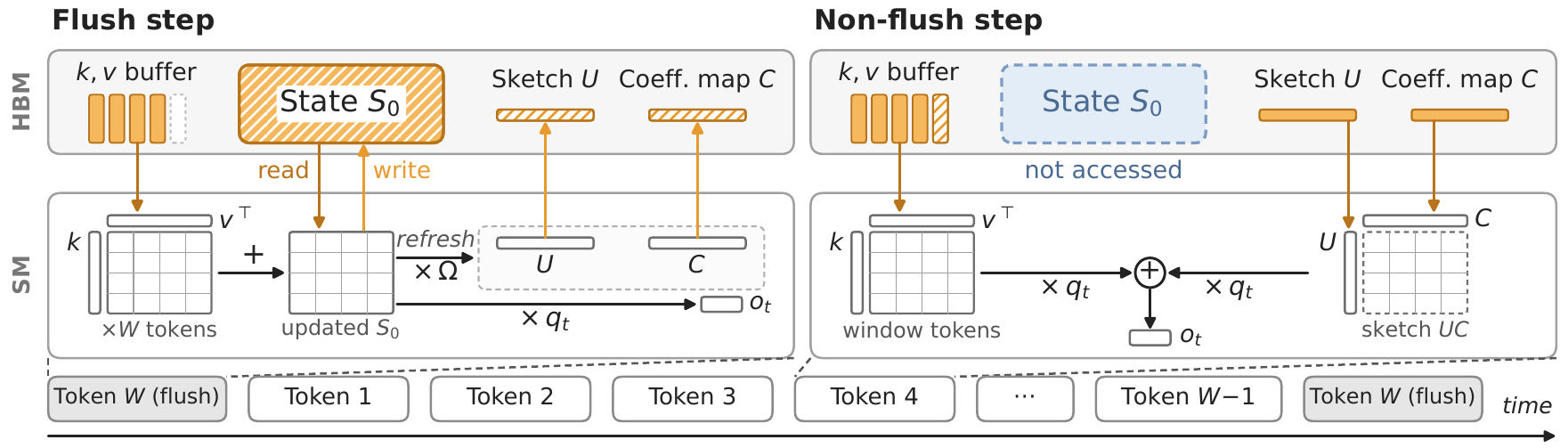}
\caption{
SketchSSM over a window.
At a flush step, SketchSSM updates the full state $S_0$
and refreshes the sketch $U$ and coefficient map $C$.
At non-flush steps, it combines $U$ and $C$ with the query
to reconstruct the output.
Note that $S_0$ is not accessed between state updates.
Algorithm~\ref{alg:sketchssm} summarizes the procedure.
}
\label{fig:sketch-overview}
\end{figure}

\name{} provides an accurate, inference-efficient solution to
Equation~\ref{eq:state_weighted_query}, as illustrated in
Fig.~\ref{fig:sketch-overview}.
To avoid the computational overhead of repeatedly computing
the query basis, we calibrate the sketching matrix
$\sketchingmatrix$ offline and fix it throughout inference.
In contrast, we refresh the sketch $\sketch$ at each window
boundary to reflect the current state, precomputing the
outputs for these fixed query basis vectors.
Finally, at each decode step between state updates,
we compute query-dependent coefficients $\coefficientvector$
that linearly combine the sketch vectors to accurately
approximate the current query's output.
We first detail these three computations, then describe how we allocate sketch ranks across heads for better accuracy. Algorithm~\ref{alg:sketchssm} (Appendix~\ref{app:sketchssm-algorithm}) summarizes the complete procedure.

\subsection{Constructing the Sketch}
\label{sec:sketch-def}

\para{Offline: Sketching Matrix ($\sketchingmatrix$)}
The columns of $(\state\state^\top)^{1/2}\sketchingmatrix$
span the low-dimensional subspace used to approximate
the state-weighted query
$(\state\state^\top)^{1/2}\effectivequery$.
The optimal $\sketchingmatrix$ therefore depends on
$\state$, which varies across requests and decode steps.
However, optimizing $\sketchingmatrix$ for each $\state$
is computationally expensive.
We therefore fix $\sketchingmatrix$ throughout inference
and calibrate it offline for each state head using the average
state geometry over the calibration distribution $\mathcal D$,
$E_0:=\mathbb E_{\mathcal D}[\state\state^\top]$.
This means that, at runtime, we still approximate
$(\state\state^\top)^{1/2}\effectivequery$ with
$(\state\state^\top)^{1/2}\sketchingmatrix\coefficientvector$,
but use a shared $\sketchingmatrix$ selected from calibration
samples using $E_0$.
The corresponding offline loss is
\begin{equation}
\mathcal L_G(\sketchingmatrix)
:=
\mathbb E_{\mathcal D}\!\left[
\min_{\coefficientvector}
\left\|
E_0^{1/2}\effectivequery
-
E_0^{1/2}\sketchingmatrix\coefficientvector
\right\|_2^2
\right].
\label{eq:m-subspace-objective}
\end{equation}
It is solved by uncentered PCA
of the state-weighted query
$z_t:=E_0^{1/2}\effectivequery$.
With $C_z:=\mathbb E_{\mathcal D}[z_tz_t^\top]$,
the solution is
$\sketchingmatrix^\star=E_0^{-1/2}P_G
\in\mathbb R^{K\times G}$,
where $P_G$ contains the leading $G$ eigenvectors of $C_z$.
Appendix~\ref{app:offline-sketch-proof}
gives the full derivation, and Appendix~\ref{app:evaluation-setup} describes the detailed calibration procedure.

\para{Per-window: Sketch ($\sketch$)}
At the flush step of each window, the buffered updates
are applied to the full state, producing $\state$
for the next window.
As described in Section~\ref{sec:mot-obs}, we precompute
the output vectors that $\state$ produces for the query
basis vectors in $\sketchingmatrix$, forming
the sketch $\sketch=\state^\top\sketchingmatrix$.
Constructing the sketch requires a single full-state read,
so we fuse this computation into the flush kernel that
updates $\state$.
The kernel constructs $\sketch$ before writing the updated
state back to memory, eliminating additional full-state traffic.

\label{sec:sketch-coeff}

\para{Per-step: Coefficient Vector ($\coefficientvector$)}
At each decode step between flushes, \name{} combines
the precomputed output vectors in $\sketch$ to approximate
$\stateoutput$ as $\approxstateoutput=\sketch\coefficientvector$.
We choose the coefficients to minimize output error
for the current state and query:
\begin{equation}
\coefficientvector^\star
:=\argmin_{c\in\mathbb R^G}
\left\|\state^\top\effectivequery-\sketch c\right\|_2^2
=\coefficientmap\effectivequery,
\qquad
\coefficientmap
:=\big(\sketch^\top\sketch\big)^\dagger
\sketch^\top\state^\top.
\label{eq:m-coeff}
\end{equation}
Appendix~\ref{app:sketch-problem} gives the derivation.
The coefficient map $\coefficientmap$ depends on the current
state, which remains fixed within a window.
We therefore compute $\coefficientmap$ once at each flush
and obtain $\coefficientvector=\coefficientmap\effectivequery$
for each subsequent query.

A cheaper alternative is to precompute a fixed map
using the average state metric $E_0$:
\begin{equation}
\coefficientvector^{\mathrm{offline}}
:=\argmin_{c\in\mathbb R^G}
\left\|
E_0^{1/2}\effectivequery-E_0^{1/2}\sketchingmatrix c
\right\|_2^2
=C_{\mathrm{offline}}\effectivequery,
\qquad
C_{\mathrm{offline}}:=
(\sketchingmatrix^\top E_0\sketchingmatrix)^\dagger
\sketchingmatrix^\top E_0.
\end{equation}
This avoids recomputing the map at each flush, but does not
adapt to the current state and substantially reduces accuracy
in our ablation (Table~\ref{tab:allocation-ablation}).
We therefore retain the state-dependent map and reduce
its computation cost using a low-rank-plus-diagonal
approximation detailed in Appendix~\ref{app:pivot-coeff}.
Table~\ref{tab:coefficient-flush-latency} shows that this
approximation accelerates the complete flush step by
$4.51\times$ relative to exact coefficient-map computation
on Nemotron 3 Super at $\bar G=8$ and batch size 256.
With this optimization, Fig.~\ref{fig:super-kernel-speedup}(b)
shows that constructing the sketch and coefficient map adds
20.3--27.1\% to flush-step latency on Nemotron 3 Super relative to the same CUDA
flush implementation without these operations.

\subsection{Allocating Sketch Ranks Across Layers and Heads}
\label{sec:sketch-alloc}

\name{} uses offline calibration to allocate a sketch-rank budget,
which determines sketch size, non-uniformly across state heads.
It assigns larger ranks to heads whose low-rank sketches leave
larger $\stateoutput$ reconstruction errors and whose
$\stateoutput$ has a stronger influence on the loss.
For each head and candidate rank $G$, let
$\delta^o_{t,l,h}(G)\in\mathbb{R}^{V}$ be the output reconstruction
error from using a rank-$G$ sketch instead of the full state,
and let $g_{t,l,h}:=\nabla_{\stateoutput[t,l,h]}\ell
\in\mathbb{R}^{V}$ be the loss gradient, which measures how
sensitive the model loss is to changes in $\stateoutput[t,l,h]$.
We define
\begin{equation}
J_{l,h}(G):=\mathbb E_t\!\left[
\big(g_{t,l,h}^{\top}\delta^o_{t,l,h}(G)\big)^2\right].
\label{eq:m-head-objective}
\end{equation}
Our goal is to minimize the sum of these scores across state heads
under a fixed mean sketch-rank budget, $\bar G$.
We solve this problem as a multiple-choice knapsack problem
using a Lagrangian relaxation, as detailed in
Appendix~\ref{app:alloc-problem}.
An offline calibration run computes all per-head candidate scores,
which we reuse to generate allocations for different mean
sketch-rank budgets $\bar G$.
Table~\ref{tab:allocation-ablation} shows that our allocation improves
accuracy over uniform allocation by \(10.04\) percentage points on average.

\FloatBarrier

\begin{figure}[!t]
\centering
\includegraphics[width=\textwidth]{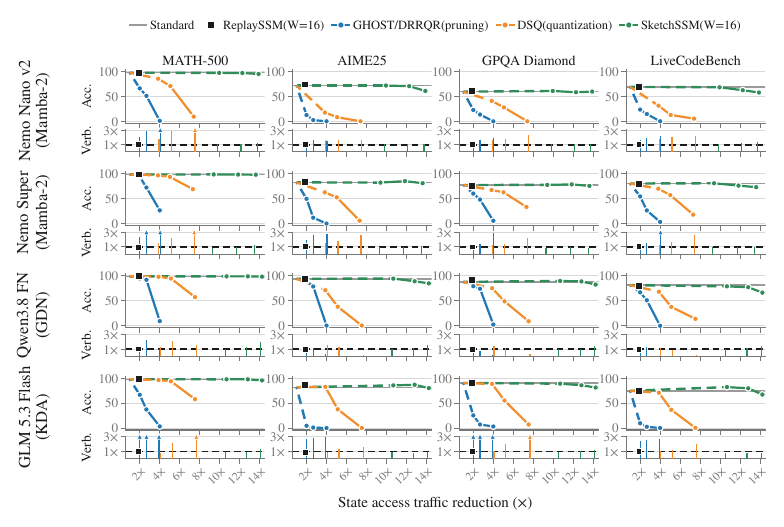}
\caption{Accuracy and verbosity versus state-access traffic
reduction relative to Standard as detailed in Appendix~\ref{sec:sketch-traffic}.
ReplaySSM and SketchSSM use $W=16$.
DSQ uses $\{8,6,4\}$-bit states;
GHOST (Nano/Super/GLM) and DRRQR (Qwen) prune
$\{50,62.5,75\}\%$ of the state.
SketchSSM uses $\bar G\in\{8,4,2\}$ for all four models.
At $\bar G=8$, which reduces state-access traffic by 9.4--10.7$\times$,
SketchSSM preserves average accuracy across the four benchmarks for all
four models, whereas pruning and quantization degrade accuracy at
smaller reductions.
Verbosity is the mean generation length normalized to Standard.
Raw accuracy results are provided in Appendix~\ref{app:decode-results}.
}
\label{fig:pareto}
\end{figure}

\section{Evaluation}
\label{sec:evaluation}

\para{Evaluation Setup}
We evaluate four models spanning three linear-attention methods:
Nemotron Nano 9B v2 and Nemotron 3 Super (Mamba-2),
Qwen3.8 Flash-Next (GDN), and GLM 5.3 Flash (KDA), all with NVFP4 weights.
We evaluate accuracy on MATH-500~\citep{math500}, AIME25~\citep{aime25}, GPQA Diamond~\citep{gpqa},
and LiveCodeBench~\citep{lcb} against state quantization, state pruning,
and full-state baselines with Standard execution and ReplaySSM~\citep{daolab2026replayssm}.
For Standard, ReplaySSM, and SketchSSM, we report pass@1 averaged over
eight seeds for AIME25 and three seeds for the other benchmarks.
SketchSSM is calibrated only on general-domain text from WikiText-2~\citep{wiki2},
which is out of domain for every evaluated benchmark; no data from the evaluated benchmarks is used to calibrate the sketching
matrices or to allocate sketch ranks.
ReplaySSM represents buffered-update baselines such as KVBuffer~\citep{zou2026kvbufferioawareservinglinear}.
For state quantization, we use Decoupled Scale Quantization (DSQ) from Q-Mamba~\citep{q-mamba}, which dynamically computes
separate scales along the two dimensions of the state.
For state pruning, we use GHOST~\citep{menezes2026ghost}
for Mamba-2- and KDA-based models and
DRRQR~\citep{nazari2026keystatereductionlinear}
for GDN-based models.
We integrate our optimized SketchSSM kernels into vLLM 0.30.0~\citep{vllm}
and conduct all evaluations on vLLM.
For speed evaluation, we measure linear-attention kernel latency on Mamba-2, GDN, and KDA, and measure end-to-end decode throughput on Nemotron 3 Super using a single NVIDIA B300 GPU.
Detailed evaluation settings are provided in
Appendix~\ref{app:evaluation-setup}.

\subsection{Accuracy}
\label{sec:acc_evaluation}

\para{Decode Benchmark Accuracy}
As shown in Fig.~\ref{fig:pareto}, at $\bar G=8$, which reduces
state-access traffic by 9.4--10.7$\times$,
SketchSSM preserves the average accuracy of the FP32 full-state baseline (Standard)
across four decode benchmarks for all four models.
At $\bar G=4$ (11.8--12.8$\times$), average accuracy decreases by at most
2.4 percentage points relative to Standard,
while $\bar G=2$ (13.6--14.2$\times$) reduces it by 2.8--7.5 points,
illustrating the tradeoff between accuracy and state-access traffic.
In contrast, at approximately 4$\times$ reductions,
8-bit state quantization and 75\% state pruning reduce
average accuracy by 1.6--30.6 and 74.1--87.6
points, respectively, relative to the full-state baseline.
Appendix~\ref{sec:sketch-traffic} defines the state-access traffic
metric, including both reads and writes, and Appendix~\ref{app:decode-results} provides
detailed results of accuracy evaluation.

\para{Recall Accuracy}
Table~\ref{tab:recall-summary} reports recall on four RULER retrieval tasks
at 16K context.
At $\bar G\in\{8,4\}$, SketchSSM stays within 0.4 percentage points of
Standard's average recall on every model (97.92--100\%),
whereas 75\% state pruning reduces it to 43.05--76.70\%.
Because SketchSSM updates the full state exactly and approximates only its reads,
read errors can propagate through activations but do not accumulate in the state
through updates, so the state retains the information written from the long context;
pruning, in contrast, removes state coordinates and the facts they hold.
Appendix~\ref{app:recall-results} details the setup and per-task results.

\begin{table}[!htb]
\centering
\small
\caption{Average recall (\%) on four RULER retrieval tasks at 16K context.
Traffic is the state-access traffic reduction relative to Standard.
State pruning uses GHOST, except DRRQR for Qwen3.8 Flash-Next.
Per-task results are in Appendix~\ref{app:recall-results}.}
\label{tab:recall-summary}
\setlength{\tabcolsep}{1.5pt}
\begin{tabular*}{\textwidth}{@{\extracolsep{\fill}}llrrrrr@{}}
\toprule
Method & Setting & Traffic & Nemotron Nano & Nemotron Super & Qwen3.8 Flash-Next & GLM 5.3 Flash \\
\midrule
\textbf{Standard} & \textbf{full-state} & {\boldmath$1.00\times$} & \textbf{98.29} & \textbf{100.00} & \textbf{100.00} & \textbf{99.40} \\
ReplaySSM & full-state & $1.88\times$ & 98.25 & 100.00 & 100.00 & 99.46 \\
\midrule
\textbf{SketchSSM} & {\boldmath$\bar G=8$} & {\boldmath$9.4$--$10.7\times$} & \textbf{98.03} & \textbf{100.00} & \textbf{100.00} & \textbf{99.74} \\
SketchSSM & $\bar G=4$ & $11.8$--$12.8\times$ & 97.92 & 100.00 & 100.00 & 99.70 \\
\midrule
State pruning & 50\% & $2.00\times$ & 93.65 & 97.36 & 99.89 & 95.23 \\
State pruning & 62.5\% & $2.67\times$ & 82.40 & 83.26 & 98.59 & 91.14 \\
State pruning & 75\% & $4.00\times$ & 76.70 & 53.03 & 43.05 & 46.78 \\
\bottomrule
\end{tabular*}
\end{table}

\begin{table}[t]
\centering
\scriptsize
\caption{Ablation of internal design choices on
Nemotron Nano 9B v2, covering sketch-basis calibration, state-dependent coefficients, and sketch-rank allocation, at $\bar G=4$. Pass@1 is averaged over eight seeds for AIME25 and three seeds for the other benchmarks.}
\label{tab:allocation-ablation}
\setlength{\tabcolsep}{1.5pt}
\begin{tabular*}{\textwidth}{@{\extracolsep{\fill}}llrrrrr@{}}
\toprule
Ablation & Method & MATH-500 & AIME25 & GPQA-D & LCB & Average \\
\midrule
Full method & SketchSSM & 96.73 & 70.42 & 58.25 & 62.33 & 71.93 \\
\midrule
Sketch basis $\sketchingmatrix$ & Random orthogonal & 94.73 ($-2.00$) & 53.33 ($-17.08$) & 53.54 ($-4.71$) & 53.23 ($-9.10$) & 63.71 ($-8.22$) \\
\midrule
Coefficient vector $\coefficientvector$ & Offline coefficient map & 90.87 ($-5.87$) & 42.08 ($-28.33$) & 53.20 ($-5.05$) & 27.72 ($-34.60$) & 53.47 ($-18.46$) \\
\midrule
\multirow{3}{*}{Sketch-rank allocation} & Uniform allocation & 93.13 ($-3.60$) & 48.33 ($-22.08$) & 55.22 ($-3.03$) & 50.90 ($-11.43$) & 61.90 ($-10.04$) \\
 & Error-only score & 94.13 ($-2.60$) & 59.17 ($-11.25$) & 57.58 ($-0.67$) & 54.39 ($-7.94$) & 66.32 ($-5.61$) \\
 & Gradient-only score & 90.53 ($-6.20$) & 35.83 ($-34.58$) & 51.68 ($-6.57$) & 44.76 ($-17.57$) & 55.70 ($-16.23$) \\
\bottomrule
\end{tabular*}
\end{table}

\para{Accuracy Ablation}
Table~\ref{tab:allocation-ablation} shows the effects of basis calibration,
state-dependent coefficients, and head-level rank allocation
at the same mean sketch rank.
First, calibrating $\sketchingmatrix$ for state-weighted queries
improves accuracy by 2.0--17.1 percentage points over
a random orthogonal basis.
Second, state-dependent coefficient maps improve accuracy
by 5.1--34.6 points over an offline map.
Finally, allocating ranks using both output error and loss sensitivity
improves accuracy by 3.0--22.1 points over uniform allocation
and yields higher average accuracy than using either criterion alone.
Fig.~\ref{fig:sketch-alloc} in Appendix~\ref{app:alloc-problem} illustrates their variation across heads,
motivating head-level allocation.
Appendix~\ref{app:read-only-comparison} shows that preserving
full-state updates while approximating reads substantially improves
accuracy by preventing propagation of state-compression errors, with sketches further improving the accuracy--traffic
tradeoff over other read representations.

\begin{figure*}[!t]
\centering
\includegraphics[width=\textwidth]{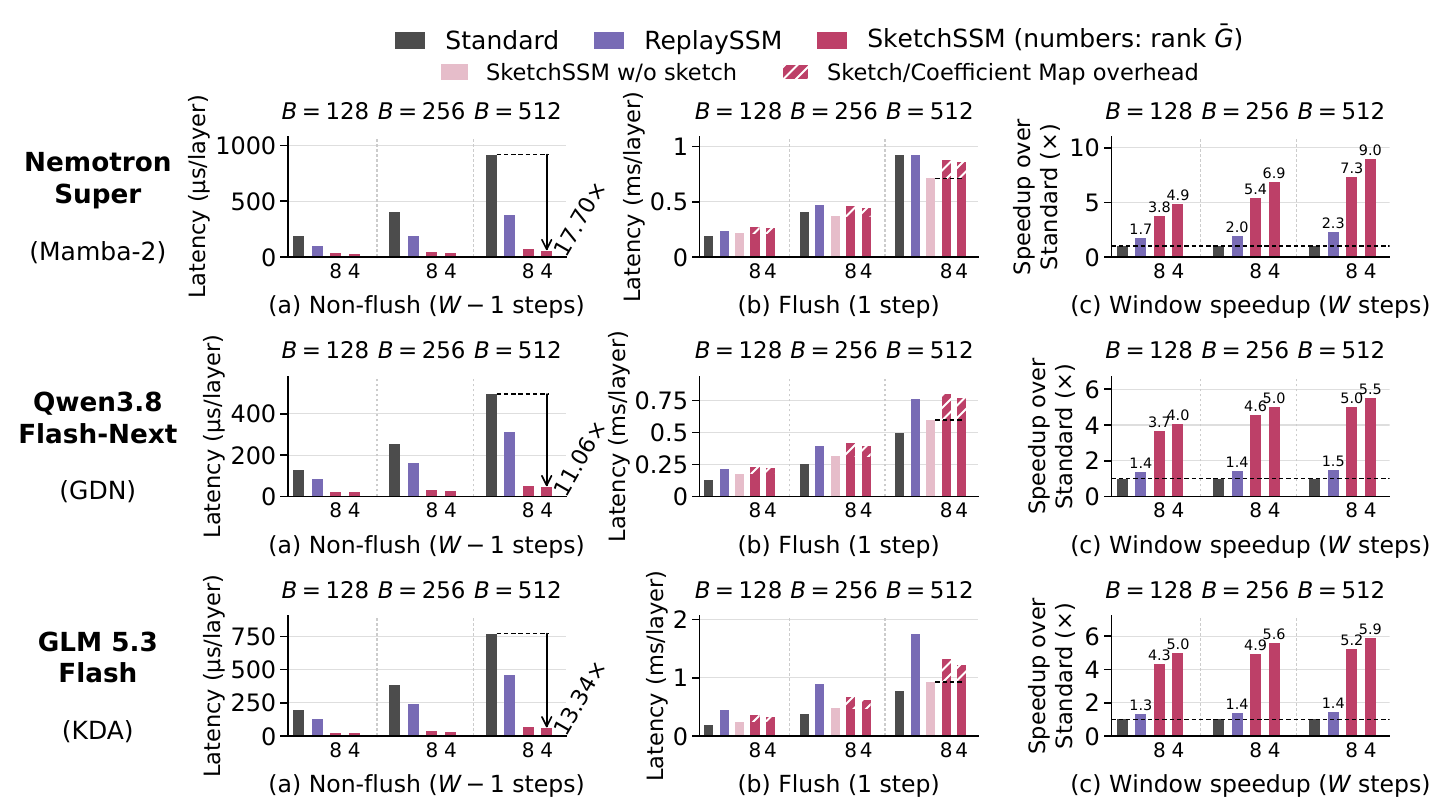}
\caption{Linear-attention latency on one NVIDIA B300 with $W=16$
and batch sizes 128, 256, and 512.
Rows show Nemotron 3 Super, Qwen3.8 Flash-Next, and
GLM 5.3 Flash, with SketchSSM at $\bar G\in\{8,4\}$.
(a) Non-flush-step latency, incurred at each of the $W-1$
steps between flush steps.
(b) Flush-step latency, incurred once per window. Hatching indicates sketch and coefficient-map construction overhead at the flush step.
(c) Total linear-attention speedup over Standard across the $W$ steps of a window.}
\label{fig:super-kernel-speedup}
\end{figure*}

\subsection{Speedups}
\label{eval:speedup}

\para{Kernel Speedups}
We evaluate Mamba-2, GDN, and KDA kernels on one NVIDIA B300
with $W=16$ and batch sizes 128, 256, and 512
(Fig.~\ref{fig:super-kernel-speedup}).
SketchSSM reduces non-flush latency by replacing full-state
reads with sketch reads.
Its fused flush kernel constructs the sketch and coefficient
map while applying state updates, limiting the additional
construction overhead.
On Nemotron 3 Super at $\bar G\in\{8,4\}$, this construction adds
20.3--27.1\% latency over the same flush kernel without it.
Including all steps in the window, total linear-attention speedups over Standard reach 7.30$\times$, 5.02$\times$, and 5.24$\times$ at $\bar G=8$ (8.99$\times$, 5.50$\times$, and 5.90$\times$ at $\bar G=4$) for Nemotron 3 Super, Qwen3.8 Flash-Next, and GLM 5.3 Flash, respectively.
ReplaySSM amortizes state updates but still reads the full state at every step,
so at batch size 512 SketchSSM is 3.24$\times$, 3.43$\times$, and 3.69$\times$
faster than ReplaySSM at $\bar G=8$ (3.98$\times$, 3.75$\times$, and 4.16$\times$ at $\bar G=4$).

\begin{figure*}[!t]
\centering
\includegraphics[width=\textwidth]{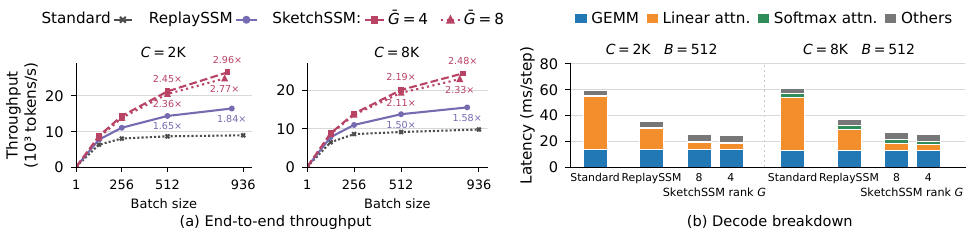}
\caption{Decoding performance for Nemotron 3 Super on one NVIDIA B300, comparing Standard, ReplaySSM,
and SketchSSM at $\bar G\in\{8,4\}$.
(a) Decode throughput at 2K and 8K up to each method's maximum
batch, averaged over two 128-step runs; labels give the speedup over Standard
at batch 512 and at each method's maximum batch.
(b) Decode latency breakdown at batch size 512.}
\label{fig:super-e2e-speedup}
\end{figure*}

\para{End-to-End Decode Speedups}
With $\bar G=8$, SketchSSM improves maximum decode throughput
by 2.77$\times$ at 2K and 2.33$\times$ at 8K
(2.96$\times$ and 2.48$\times$ at $\bar G=4$) over the
Standard full-state baseline.
Fig.~\ref{fig:super-e2e-speedup}(a) shows throughput
across batch sizes, up to each method's capacity limit.
These gains come from reducing recurrent computation,
which accounts for much of the baseline's decode time.
At batch size 512 with $\bar G=8$, linear attention's share of decode time
falls from 68.5\% to 22.3\% at 2K and from 67.1\% to 21.1\%
at 8K (Fig.~\ref{fig:super-e2e-speedup}(b)),
so linear attention no longer dominates decode time.

\section{Conclusion}

We present \name{}, which preserves exact full-state updates while
using compact sketches for output reconstruction.
We construct the sketch at runtime from the current state using
an offline-calibrated sketching matrix.
Sketch ranks are also allocated offline based on each state head's
output error and its impact on model loss.
Across four hybrid-attention models, \name{} at a mean sketch rank of 8 reduces state-access traffic by approximately $10\times$ while matching the average accuracy of the FP32 full-state baseline. At this rank on one NVIDIA B300, it achieves $5.02$--$7.30\times$ linear-attention speedups across Mamba-2, GDN, and KDA, yielding up to $2.77\times$ higher maximum throughput over the full-state baseline.

\clearpage

\clearpage
\bibliography{references}

@article{woodruff2014sketching,
    author = {Woodruff, David P.},
    title = {Sketching as a Tool for Numerical Linear Algebra},
    journal = {Foundations and Trends in Theoretical Computer Science},
    volume = {10},
    number = {1-2},
    pages = {1-157},
    year = {2014},
    month = {10},
    issn = {1551-305X},
    doi = {10.1561/0400000060},
    url = {https://doi.org/10.1561/0400000060},
    eprint = {https://www.emerald.com/fttcs/article-pdf/10/1-2/1/11485699/0400000060en.pdf},
}

@article{halko2011finding,
author = {Halko, N. and Martinsson, P. G. and Tropp, J. A.},
title = {Finding Structure with Randomness: Probabilistic Algorithms for Constructing Approximate Matrix Decompositions},
journal = {SIAM Review},
volume = {53},
number = {2},
pages = {217-288},
year = {2011},
doi = {10.1137/090771806},

URL = { 
    
        https://doi.org/10.1137/090771806
    
    

},
eprint = { 
    
        https://doi.org/10.1137/090771806
    
    

}
,
}

@inproceedings{indyk2019learning,
 author = {Indyk, Piotr and Vakilian, Ali and Yuan, Yang},
 booktitle = {Advances in Neural Information Processing Systems},
 editor = {H. Wallach and H. Larochelle and A. Beygelzimer and F. d\textquotesingle Alch\'{e}-Buc and E. Fox and R. Garnett},
 pages = {},
 publisher = {Curran Associates, Inc.},
 title = {Learning-Based Low-Rank Approximations},
 url = {https://proceedings.neurips.cc/paper_files/paper/2019/file/1625abb8e458a79765c62009235e9d5b-Paper.pdf},
 volume = {32},
 year = {2019}
}

@inproceedings{liberty2013simple,
author = {Liberty, Edo},
title = {Simple and deterministic matrix sketching},
year = {2013},
isbn = {9781450321747},
publisher = {Association for Computing Machinery},
address = {New York, NY, USA},
url = {https://doi.org/10.1145/2487575.2487623},
doi = {10.1145/2487575.2487623},
booktitle = {Proceedings of the 19th ACM SIGKDD International Conference on Knowledge Discovery and Data Mining},
pages = {581–588},
numpages = {8},
location = {Chicago, Illinois, USA},
series = {KDD '13}
}

@InProceedings{katharopoulos2020transformers,
  title = 	 {Transformers are {RNN}s: Fast Autoregressive Transformers with Linear Attention},
  author =       {Katharopoulos, Angelos and Vyas, Apoorv and Pappas, Nikolaos and Fleuret, Fran{\c{c}}ois},
  booktitle = 	 {Proceedings of the 37th International Conference on Machine Learning},
  pages = 	 {5156--5165},
  year = 	 {2020},
  editor = 	 {III, Hal Daumé and Singh, Aarti},
  volume = 	 {119},
  series = 	 {Proceedings of Machine Learning Research},
  month = 	 {13--18 Jul},
  publisher =    {PMLR},
  url = 	 {https://proceedings.mlr.press/v119/katharopoulos20a.html}
}

@InProceedings{dao2024transformers,
  title = 	 {Transformers are {SSM}s: Generalized Models and Efficient Algorithms Through Structured State Space Duality},
  author =       {Dao, Tri and Gu, Albert},
  booktitle = 	 {Proceedings of the 41st International Conference on Machine Learning},
  pages = 	 {10041--10071},
  year = 	 {2024},
  editor = 	 {Salakhutdinov, Ruslan and Kolter, Zico and Heller, Katherine and Weller, Adrian and Oliver, Nuria and Scarlett, Jonathan and Berkenkamp, Felix},
  volume = 	 {235},
  series = 	 {Proceedings of Machine Learning Research},
  month = 	 {21--27 Jul},
  publisher =    {PMLR},
  url = 	 {https://proceedings.mlr.press/v235/dao24a.html}
}

@inproceedings{yang2025gated,
title={Gated Delta Networks: Improving Mamba2 with Delta Rule},
author={Songlin Yang and Jan Kautz and Ali Hatamizadeh},
booktitle={The Thirteenth International Conference on Learning Representations},
year={2025},
url={https://openreview.net/forum?id=r8H7xhYPwz}
}

@misc{kimiteam2025kimi,
    title         = {Kimi Linear: An Expressive, Efficient Attention Architecture},
    author        = {Zhang, Yu  and Lin, Zongyu  and Yao, Xingcheng  and Hu, Jiaxi  and Meng, Fanqing  and Liu, Chengyin  and Men, Xin  and Yang, Songlin  and Li, Zhiyuan  and Li, Wentao  and Lu, Enzhe  and Liu, Weizhou  and Chen, Yanru  and Xu, Weixin  and Yu, Longhui  and Wang, Yejie  and Fan, Yu  and Zhong, Longguang  and Yuan, Enming  and Zhang, Dehao  and Zhang, Yizhi  and T. Liu, Y.  and Wang, Haiming  and Fang, Shengjun  and He, Weiran  and Liu, Shaowei  and Li, Yiwei  and Su, Jianlin  and Qiu, Jiezhong  and Pang, Bo  and Yan, Junjie  and Jiang, Zhejun  and Huang, Weixiao  and Yin, Bohong  and You, Jiacheng  and Wei, Chu  and Wang, Zhengtao  and Hong, Chao  and Chen, Yutian  and Chen, Guanduo  and Wang, Yucheng  and Zheng, Huabin  and Wang, Feng  and Liu, Yibo  and Dong, Mengnan  and Zhang, Zheng  and Pan, Siyuan  and Wu, Wenhao  and Wu, Yuhao  and Guan, Longyu  and Tao, Jiawen  and Fu, Guohong  and Xu, Xinran  and Wang, Yuzhi  and Lai, Guokun  and Wu, Yuxin  and Zhou, Xinyu  and Yang, Zhilin  and Du, Yulun},
    year          = {2025},
    eprint        = {2510.26692},
    archivePrefix = {arXiv},
    primaryClass  = {cs.CL}
}

@misc{lieber2024jamba,
      title={Jamba: A Hybrid Transformer-Mamba Language Model}, 
      author={Opher Lieber and Barak Lenz and Hofit Bata and Gal Cohen and Jhonathan Osin and Itay Dalmedigos and Erez Safahi and Shaked Meirom and Yonatan Belinkov and Shai Shalev-Shwartz and Omri Abend and Raz Alon and Tomer Asida and Amir Bergman and Roman Glozman and Michael Gokhman and Avashalom Manevich and Nir Ratner and Noam Rozen and Erez Shwartz and Mor Zusman and Yoav Shoham},
      year={2024},
      eprint={2403.19887},
      archivePrefix={arXiv},
      primaryClass={cs.CL},
      url={https://arxiv.org/abs/2403.19887}, 
}

@misc{nvidia2026nemotron3super,
      title={Nemotron 3 Super: Open, Efficient Mixture-of-Experts Hybrid Mamba-Transformer Model for Agentic Reasoning}, 
      author={NVIDIA Team},
      year={2026},
      eprint={2604.12374},
      archivePrefix={arXiv},
      primaryClass={cs.LG},
      url={https://arxiv.org/abs/2604.12374}, 
}

@techreport{qwen2026design,
    title       = {On the Design of {Qwen3.8-Next} Architecture: Evaluation, Efficiency, and Training Stability},
    author      = {{Qwen Team}},
    institution = {Alibaba Group},
    month       = {August},
    year        = {2026}
}

@misc{qwen3.8flashnext,
    title  = {{Qwen3.8-Flash-Next}: A New Architecture, Towards Ultimate Cost-Efficiency},
    author = {{Qwen Team}},
    month  = {August},
    year   = {2026},
    url    = {https://qwen.ai/blog?id=qwen3.8-flash-next}
}

@misc{glm5team2026glm5vibecodingagentic,
      title={GLM-5: from Vibe Coding to Agentic Engineering}, 
      author={GLM-5-Team},
      year={2026},
      eprint={2602.15763},
      archivePrefix={arXiv},
      primaryClass={cs.LG},
      url={https://arxiv.org/abs/2602.15763}, 
}

@inproceedings{menezes2026ghost,
  title     = {{GHOST}: Grouped Hidden-state Output-aware Selection and Truncation},
  author    = {Menezes, Michael and Kyrillidis, Anastasios},
  booktitle = {Proceedings of the International Conference on Machine Learning (ICML)},
  year      = {2026}
}

@inproceedings{q-mamba,
    title = "{Q}-Mamba: Towards more efficient Mamba models via post-training quantization",
    author = "Tianqi, Chen  and
      Chen, Yuanteng  and
      Wang, Peisong  and
      Xu, Weixiang  and
      Zhu, Zeyu  and
      Cheng, Jian",
    editor = "Che, Wanxiang  and
      Nabende, Joyce  and
      Shutova, Ekaterina  and
      Pilehvar, Mohammad Taher",
    booktitle = "Findings of the Association for Computational Linguistics: ACL 2025",
    month = jul,
    year = "2025",
    address = "Vienna, Austria",
    publisher = "Association for Computational Linguistics",
    url = "https://aclanthology.org/2025.findings-acl.551/",
    doi = "10.18653/v1/2025.findings-acl.551",
    pages = "10594--10610",
    ISBN = "979-8-89176-256-5"
}

@misc{daolab2026replayssm,
  author       = {Liou, Ze-Wei and Dao, Tri},
  title        = {{ReplaySSM}: Cache {SSM} Inputs, Not State},
  year         = {2026},
  month        = jun,
  howpublished = {Dao AI Lab Blog},
  url          = {https://dao-lab.ai/blog/2026/replayssm/},
  note         = {Accessed: 2026-09-07}
}

@misc{nazari2026keystatereductionlinear,
      title={The Key to State Reduction in Linear Attention: A Rank-based Perspective}, 
      author={Philipp Nazari and T. Konstantin Rusch},
      year={2026},
      eprint={2602.04852},
      archivePrefix={arXiv},
      primaryClass={cs.LG},
      url={https://arxiv.org/abs/2602.04852}, 
}

@InProceedings{chiang2025quamba2,
  title = 	 {Quamba2: A Robust and Scalable Post-training Quantization Framework for Selective State Space Models},
  author =       {Chiang, Hung-Yueh and Chang, Chi-Chih and Frumkin, Natalia and Wu, Kai-Chiang and Abdelfattah, Mohamed S. and Marculescu, Diana},
  booktitle = 	 {Proceedings of the 42nd International Conference on Machine Learning},
  pages = 	 {10411--10427},
  year = 	 {2025},
  editor = 	 {Singh, Aarti and Fazel, Maryam and Hsu, Daniel and Lacoste-Julien, Simon and Berkenkamp, Felix and Maharaj, Tegan and Wagstaff, Kiri and Zhu, Jerry},
  volume = 	 {267},
  series = 	 {Proceedings of Machine Learning Research},
  month = 	 {13--19 Jul},
  publisher =    {PMLR},
  url = 	 {https://proceedings.mlr.press/v267/chiang25a.html}
}

@InProceedings{arora2024simple,
  title = 	 {Simple linear attention language models balance the recall-throughput tradeoff},
  author =       {Arora, Simran and Eyuboglu, Sabri and Zhang, Michael and Timalsina, Aman and Alberti, Silas and Zou, James and Rudra, Atri and Re, Christopher},
  booktitle = 	 {Proceedings of the 41st International Conference on Machine Learning},
  pages = 	 {1763--1840},
  year = 	 {2024},
  editor = 	 {Salakhutdinov, Ruslan and Kolter, Zico and Heller, Katherine and Weller, Adrian and Oliver, Nuria and Scarlett, Jonathan and Berkenkamp, Felix},
  volume = 	 {235},
  series = 	 {Proceedings of Machine Learning Research},
  month = 	 {21--27 Jul},
  publisher =    {PMLR},
  url = 	 {https://proceedings.mlr.press/v235/arora24a.html}
}

@inproceedings{
qin2024hgrn2,
title={{HGRN}2: Gated Linear {RNN}s with State Expansion},
author={Zhen Qin and Songlin Yang and Weixuan Sun and Xuyang Shen and Dong Li and Weigao Sun and Yiran Zhong},
booktitle={First Conference on Language Modeling},
year={2024},
url={https://openreview.net/forum?id=y6SqbJfCSk}
}

@inproceedings{
pan2025scalinglinearattentionsparse,
title={Scaling Linear Attention Capacity with Sparse State Expansion},
author={Yuqi Pan and Yongqi An and Zheng Li and Yuhong Chou and Rui-Jie Zhu and Xiaohui Wang and Mingxuan Wang and Jinqiao Wang and Guoqi Li},
booktitle={The Fourteenth International Conference on Learning Representations},
year={2026},
url={https://openreview.net/forum?id=R6DrJ4tnGV}
}

@inproceedings{saxena2024eigenattention,
    title = "Eigen Attention: Attention in Low-Rank Space for {KV} Cache Compression",
    author = "Saxena, Utkarsh  and
      Saha, Gobinda  and
      Choudhary, Sakshi  and
      Roy, Kaushik",
    editor = "Al-Onaizan, Yaser  and
      Bansal, Mohit  and
      Chen, Yun-Nung",
    booktitle = "Findings of the Association for Computational Linguistics: EMNLP 2024",
    month = nov,
    year = "2024",
    address = "Miami, Florida, USA",
    publisher = "Association for Computational Linguistics",
    url = "https://aclanthology.org/2024.findings-emnlp.899/",
    doi = "10.18653/v1/2024.findings-emnlp.899",
    pages = "15332--15344"
}

@inproceedings{schlag2021linear,
      title={Linear Transformers Are Secretly Fast Weight Programmers}, 
      author={Imanol Schlag and Kazuki Irie and J\"urgen Schmidhuber},
      booktitle={Proc. Int. Conf. on Machine Learning (ICML)},
      address = {Virtual only},
      month = jul,
      year={2021}
}

@InProceedings{pmlr-v235-yang24ab,
  title = 	 {Gated Linear Attention Transformers with Hardware-Efficient Training},
  author =       {Yang, Songlin and Wang, Bailin and Shen, Yikang and Panda, Rameswar and Kim, Yoon},
  booktitle = 	 {Proceedings of the 41st International Conference on Machine Learning},
  pages = 	 {56501--56523},
  year = 	 {2024},
  editor = 	 {Salakhutdinov, Ruslan and Kolter, Zico and Heller, Katherine and Weller, Adrian and Oliver, Nuria and Scarlett, Jonathan and Berkenkamp, Felix},
  volume = 	 {235},
  series = 	 {Proceedings of Machine Learning Research},
  month = 	 {21--27 Jul},
  publisher =    {PMLR},
  url = 	 {https://proceedings.mlr.press/v235/yang24ab.html}
}

@misc{zou2026kvbufferioawareservinglinear,
      title={KVBuffer: IO-aware Serving for Linear Attention}, 
      author={Longwei Zou and Lin Zhong},
      year={2026},
      eprint={2605.19049},
      archivePrefix={arXiv},
      primaryClass={cs.LG},
      url={https://arxiv.org/abs/2605.19049}, 
}

@misc{zhang2026dampdecayawaremixedprecisionrecurrentstate,
      title={DAMP: Decay-Aware Mixed-Precision Recurrent-State Quantization}, 
      author={Tao Zhang and Jianchao Tan and Pingwei Sun and Yanqi Yu and Zixu Jiang and Yuchen Xie and Xunliang Cai and Ziqian Zeng},
      year={2026},
      eprint={2608.27513},
      archivePrefix={arXiv},
      primaryClass={cs.LG},
      url={https://arxiv.org/abs/2608.27513}, 
}

@misc{hsieh2024ruler,
      title={RULER: What's the Real Context Size of Your Long-Context Language Models?}, 
      author={Cheng-Ping Hsieh and Simeng Sun and Samuel Kriman and Shantanu Acharya and Dima Rekesh and Fei Jia and Yang Zhang and Boris Ginsburg},
      year={2024},
      eprint={2404.06654},
      archivePrefix={arXiv},
      primaryClass={cs.CL},
      url={https://arxiv.org/abs/2404.06654}, 
}

@inproceedings{yang2024parallelizing,
 author = {Yang, Songlin and Wang, Bailin and Zhang, Yu and Shen, Yikang and Kim, Yoon},
 booktitle = {Advances in Neural Information Processing Systems},
 doi = {10.52202/079017-3668},
 editor = {A. Globerson and L. Mackey and D. Belgrave and A. Fan and U. Paquet and J. Tomczak and C. Zhang},
 pages = {115491--115522},
 publisher = {Curran Associates, Inc.},
 title = {Parallelizing Linear Transformers with the Delta Rule over Sequence Length},
 url = {https://proceedings.neurips.cc/paper_files/paper/2024/file/d13a3eae72366e61dfdc7eea82eeb685-Paper-Conference.pdf},
 volume = {37},
 year = {2024}
}

@inproceedings{wang2025svdllm,
  title={{SVD}-{LLM}: Truncation-aware Singular Value Decomposition for Large Language Model Compression},
  author={Xin Wang and Yu Zheng and Zhongwei Wan and Mi Zhang},
  booktitle={International Conference on Learning Representations (ICLR)},
  year={2025},
  url={https://openreview.net/forum?id=LNYIUouhdt}
}

@misc{yuan2023asvd,
      title={ASVD: Activation-aware Singular Value Decomposition for Compressing Large Language Models}, 
      author={Zhihang Yuan and Yuzhang Shang and Yue Song and Dawei Yang and Qiang Wu and Yan Yan and Guangyu Sun},
      year={2025},
      eprint={2312.05821},
      archivePrefix={arXiv},
      primaryClass={cs.CL},
      url={https://arxiv.org/abs/2312.05821}, 
}

@article{Eckart_Young_1936, title={The Approximation of One Matrix by Another of Lower Rank}, volume={1}, DOI={10.1007/BF02288367}, number={3}, journal={Psychometrika}, author={Eckart, Carl and Young, Gale}, year={1936}, pages={211–218}}

@article{math500,
      title={Let's Verify Step by Step}, 
      author={Lightman, Hunter and Kosaraju, Vineet and Burda, Yura and Edwards, Harri and Baker, Bowen and Lee, Teddy and Leike, Jan and Schulman, John and Sutskever, Ilya and Cobbe, Karl},
      journal={arXiv preprint arXiv:2305.20050},
      year={2023}
}

@misc{aime25,
      title={American Invitational Mathematics Examination (AIME) 2025}, 
      author={Zhang, Yifan and Math-AI, Team},
      year={2025},
}

@inproceedings{gpqa,
      title={{GPQA}: A Graduate-Level Google-Proof Q\&A Benchmark},
      author={David Rein and Betty Li Hou and Asa Cooper Stickland and Jackson Petty and Richard Yuanzhe Pang and Julien Dirani and Julian Michael and Samuel R. Bowman},
      booktitle={First Conference on Language Modeling},
      year={2024},
      url={https://openreview.net/forum?id=Ti67584b98}
}

@article{lcb,
    title={LiveCodeBench: Holistic and Contamination Free Evaluation of Large Language Models for Code},
    author={Jain, Naman and Han, King and Gu, Alex and Li, Wen-Ding and Yan, Fanjia and Zhang, Tianjun and Wang, Sida and Solar-Lezama, Armando and Sen, Koushik and Stoica, Ion},
    journal={arXiv preprint arXiv:2403.07974},
    year={2024}
}

@inproceedings{vllm,
  title={Efficient Memory Management for Large Language Model Serving with PagedAttention},
  author={Woosuk Kwon and Zhuohan Li and Siyuan Zhuang and Ying Sheng and Lianmin Zheng and Cody Hao Yu and Joseph E. Gonzalez and Hao Zhang and Ion Stoica},
  booktitle={Proceedings of the ACM SIGOPS 29th Symposium on Operating Systems Principles},
  year={2023}
}

@misc{wiki2,
      title={Pointer Sentinel Mixture Models},
      author={Stephen Merity and Caiming Xiong and James Bradbury and Richard Socher},
      year={2016},
      eprint={1609.07843},
      archivePrefix={arXiv},
      primaryClass={cs.CL}
}
\bibliographystyle{references}

\appendix
\clearpage
\section{Derivation of SketchSSM}
SketchSSM solves two related approximation problems. First, it calibrates the
shared sketching matrix \(\sketchingmatrix\) from many state--query samples.
Second, for the current state and query, it computes the coefficient vector
\(\coefficientvector\) used to combine the columns of the state sketch. We
derive these two steps separately and then explain their efficient runtime
implementation.

\subsection{Derivation of the Optimal Offline Sketching Matrix (\(\sketchingmatrix\))}
\label{app:offline-sketch-proof}

We derive the PCA solution used in
Equation~\ref{eq:m-subspace-objective}. Recall that
\(E_0\in\mathbb R^{K\times K}\) is the calibration-averaged state geometry,
\(\effectivequery\in\mathbb R^K\), and
\(\sketchingmatrix\in\mathbb R^{K\times G}\).
The goal is to choose the \(G\)-dimensional column space of \(\sketchingmatrix\) that minimizes the offline reconstruction loss in Equation~\ref{eq:m-subspace-objective}.

\paragraph{Step 1: Convert the weighted error to an ordinary Euclidean error.}
Define
\begin{equation}
z_t:=E_0^{1/2}\effectivequery\in\mathbb R^K,
\qquad
B:=E_0^{1/2}\sketchingmatrix\in\mathbb R^{K\times G}.
\label{eq:app-whitened-variables}
\end{equation}
These are only changes of variables. In particular,
\begin{equation}
\left\|E_0^{1/2}
(\effectivequery-\sketchingmatrix\coefficientvector)\right\|_2^2
=\|z_t-B\coefficientvector\|_2^2.
\label{eq:app-whitened-error}
\end{equation}
Thus, for a fixed \(B\), the inner problem asks for the linear combination
of its \(G\) columns that is closest to \(z_t\).

\paragraph{Step 2: Solve for the best coefficient at one sample.}
Expanding the squared distance gives
\begin{align}
\|z_t-Bc\|_2^2
&=(z_t-Bc)^\top(z_t-Bc) \nonumber\\
&=z_t^\top z_t-2c^\top B^\top z_t+c^\top B^\top Bc.
\label{eq:app-expand-offline-ls}
\end{align}
Differentiating with respect to \(c\) and setting the derivative to zero
produces the normal equation
\begin{equation}
B^\top Bc=B^\top z_t.
\label{eq:app-offline-normal-equation}
\end{equation}
If \(B\) has full column rank, \(B^\top B\) is invertible, and hence
\begin{equation}
c_{t,\mathrm{off}}^\star(B)=(B^\top B)^{-1}B^\top z_t.
\label{eq:app-optimal-offline-coeff}
\end{equation}
The reconstructed vector is
\(Bc_{t,\mathrm{off}}^\star=\Pi_Bz_t\), where
\begin{equation}
\Pi_B:=B(B^\top B)^{-1}B^\top
\label{eq:app-projector}
\end{equation}
is the orthogonal projector onto the column space of \(B\). Therefore, the
minimum sample error is \(\|(I-\Pi_B)z_t\|_2^2\), the energy left outside
that column space. If \(B\) is rank deficient, the same statements hold with
the Moore--Penrose pseudoinverse.

\paragraph{Step 3: Average the projection error over calibration samples.}
Let \(C_z:=\mathbb E_{\mathcal D}[z_tz_t^\top]\in\mathbb R^{K\times K}\).
Because \(\Pi_B\) is symmetric and idempotent,
\((I-\Pi_B)^\top(I-\Pi_B)=I-\Pi_B\). We also use the scalar identity
\(x^\top Ax=\operatorname{tr}(Axx^\top)\). Applying these two facts one line
at a time gives
\begin{align}
\mathcal L_G(\sketchingmatrix)
&=\mathbb E_{\mathcal D}
  \left[\|(I-\Pi_B)z_t\|_2^2\right] \nonumber\\
&=\mathbb E_{\mathcal D}
  \left[z_t^\top(I-\Pi_B)z_t\right] \nonumber\\
&=\mathbb E_{\mathcal D}
  \left[\operatorname{tr}((I-\Pi_B)z_tz_t^\top)\right] \nonumber\\
&=\operatorname{tr}((I-\Pi_B)C_z) \nonumber\\
&=\operatorname{tr}(C_z)-\operatorname{tr}(\Pi_BC_z).
\label{eq:app-offline-pca}
\end{align}

\paragraph{Step 4: Choose the best \(G\)-dimensional subspace.}
The total energy \(\operatorname{tr}(C_z)\) does not depend on \(B\).
Minimizing Equation~\ref{eq:app-offline-pca} is therefore equivalent to
maximizing the captured energy \(\operatorname{tr}(\Pi_BC_z)\). Write the
eigendecomposition as
\begin{equation}
C_z=P\operatorname{Diag}(\lambda_1,\ldots,\lambda_K)P^\top,
\qquad
\lambda_1\geq\cdots\geq\lambda_K\geq0.
\label{eq:app-cz-eigendecomposition}
\end{equation}
The PCA variational principle states that a rank-\(G\) projector captures
the most energy by selecting the eigenvectors with the \(G\) largest
eigenvalues. To see this directly, let
\(Q=[q_1,\ldots,q_G]\) be any orthonormal basis for the chosen subspace, so
\(\Pi_B=QQ^\top\). Then
\begin{equation}
\operatorname{tr}(\Pi_BC_z)
=\sum_{g=1}^{G}q_g^\top C_zq_g
=\sum_{i=1}^{K}\lambda_i
  \underbrace{\sum_{g=1}^{G}(p_i^\top q_g)^2}_{w_i}.
\label{eq:app-pca-energy}
\end{equation}
The weights satisfy \(0\leq w_i\leq1\) and
\(\sum_iw_i=G\). Because the eigenvalues are sorted from largest to smallest,
the weighted sum is maximized by assigning unit weight to the first \(G\)
eigenvectors. Let \(P_G=[p_1,\ldots,p_G]\) contain those eigenvectors. We can
therefore choose \(B^\star=P_G\), for which
\(\Pi_{B^\star}=P_GP_G^\top\).

\paragraph{Step 5: Map the PCA basis back to \(\sketchingmatrix\).}
From the definition \(B=E_0^{1/2}\sketchingmatrix\), the choice
\(B^\star=P_G\) gives
\begin{equation}
\sketchingmatrix^\star=E_0^{-1/2}P_G,
\qquad
c_{t,\mathrm{off}}^\star=P_G^\top z_t,
\qquad
\min_{\sketchingmatrix}\mathcal L_G(\sketchingmatrix)
=\sum_{i=G+1}^{K}\lambda_i.
\label{eq:app-offline-solution}
\end{equation}
The final equality has a direct interpretation: PCA retains the first \(G\)
eigen-directions, so the minimum residual is the energy in the remaining
\(K-G\) directions. No centering is performed, which is why the main text
calls this procedure uncentered PCA. The coefficient
\(c_{t,\mathrm{off}}^\star\) is used only inside this offline calibration
objective; inference uses the current-state coefficient derived in
Appendix~\ref{app:sketch-problem}. If \(E_0\) is singular,
\(E_0^{-1/2}\) denotes the Moore--Penrose pseudoinverse on the support of
\(E_0\); directions in its null space do not contribute to the output error.

\subsection{Derivation of the Coefficient Vector (\(\coefficientvector\))}

\subsubsection{Derivation of the Optimal Coefficient Vector}
\label{app:sketch-problem}

We next derive the per-step coefficient used with the current state sketch.
Within one window, \(\state\in\mathbb R^{K\times V}\) and
\(\sketch\in\mathbb R^{V\times G}\) are fixed, whereas
\(\effectivequery\in\mathbb R^K\) changes at every step. The exact output
\(\stateoutput=\state^\top\effectivequery\in\mathbb R^V\) is approximated by
\(\sketch c\), a linear combination of the \(G\) columns of \(\sketch\).

\paragraph{Step 1: Write and expand the least-squares problem.}
For the current query, we solve
\begin{align}
\coefficientvector^\star
&:=\argmin_{c\in\mathbb R^G}\|\stateoutput-\sketch c\|_2^2, \nonumber\\
\|\stateoutput-\sketch c\|_2^2
&={\stateoutput}^\top\stateoutput-2c^\top\sketch^\top\stateoutput
  +c^\top\sketch^\top\sketch c.
\label{eq:app-expand-runtime-ls}
\end{align}

\paragraph{Step 2: Solve the normal equation.}
Differentiating the second line with respect to \(c\) gives
\begin{equation}
2\sketch^\top(\sketch c-\stateoutput)=0
\quad\Longrightarrow\quad
\sketch^\top\sketch c=\sketch^\top\stateoutput.
\label{eq:app-runtime-normal-equation}
\end{equation}
Substituting \(\stateoutput=\state^\top\effectivequery\) and choosing the
minimum-norm solution yields
\begin{equation}
\begin{aligned}
\coefficientvector^\star
&=(\sketch^\top\sketch)^\dagger
  \sketch^\top\state^\top\effectivequery\\
&=\coefficientmap\effectivequery,
\qquad
\coefficientmap:=(\sketch^\top\sketch)^\dagger
  \sketch^\top\state^\top\in\mathbb R^{G\times K}.
\end{aligned}
\label{eq:app-optimal-coeff}
\end{equation}
Here, \(\dagger\) denotes the Moore--Penrose pseudoinverse. The first factor
\((\sketch^\top\sketch)^\dagger\in\mathbb R^{G\times G}\) corrects for
correlations among the sketch basis vectors, and
\(\sketch^\top\state^\top\in\mathbb R^{G\times K}\) measures how the full
state maps a query into those basis vectors.

With the exact least-squares coefficient map, the reconstructed output is
\begin{equation}
\sketch\coefficientvector^\star
=\sketch(\sketch^\top\sketch)^\dagger\sketch^\top
 \state^\top\effectivequery
=\Pi_{\sketch}\stateoutput,
\label{eq:app-range-projection}
\end{equation}
where \(\Pi_{\sketch}\) is the orthogonal projector onto
\(\mathrm{range}(\sketch)=\mathrm{range}(\state^\top\sketchingmatrix)\).
This has the range-projection form of \citet{halko2011finding},
with the calibrated \(\sketchingmatrix\) serving as the test matrix.

\paragraph{Step 3: Reuse the query-independent factors.}
Both \(\state\) and \(\sketch\) remain fixed within a window, so
\(\coefficientmap\) is also fixed. SketchSSM constructs
\(\coefficientmap\) once at the flush step. Each non-flush step then needs
only the matrix--vector product
\(\coefficientvector^\star=\coefficientmap\effectivequery\). The remaining
question is how to evaluate this product without serially forming the
effective query; Appendix~\ref{app:wy-coefficients} describes that separate
execution optimization.

\subsubsection{Low-Rank-Plus-Diagonal Approximation for the Coefficient Map}
\label{app:pivot-coeff}

Constructing the exact coefficient map
$\coefficientmap=(\sketch^{\top}\sketch)^\dagger\sketch^\top\state^\top$
at every flush requires forming and factoring $\sketch^\top\sketch$ and
forming $\sketch^\top\state^\top$, which costs
$O(KVG+VG^2+G^3)$ arithmetic per head.
This section derives the approximation used to reduce that flush-step cost.

\paragraph{Step 1: Express the solve in basis-aligned coordinates.}
For the derivation, we express the coefficient solve in
coordinates aligned with the calibrated query subspace.
This change of coordinates does not require storing or
updating the recurrent state in the rotated basis.
We use an orthonormal basis $\sketchingmatrix$ for this
subspace and complete it to an orthogonal matrix $R^\top$,
where $R^\top R=I_K$. Define
\begin{equation}
S'_0:=R\state\in\mathbb R^{K\times V},
\qquad
x_t:=R\effectivequery\in\mathbb R^K,
\qquad
J:=[e_1\;\cdots\;e_G]\in\mathbb R^{K\times G}.
\label{eq:pivot-coordinates}
\end{equation}
Here, $J$ selects the first $G$ basis-aligned coordinates,
so $\sketchingmatrix=R^\top J$ and
$\sketch=\state^\top R^\top J={S'_0}^{\top}J$.
This coordinate change preserves the output reconstruction
error exactly.

Define the normalized state metric
\begin{equation}
\mu:=\|S'_0\|_F^2/K,
\qquad
H:=S'_0{S'_0}^{\top}/\mu\in\mathbb R^{K\times K},
\label{eq:pivot-state-metric}
\end{equation}
where we set $\mu=1$ for a zero state.
The normalized output error is
\begin{equation}
\|{S'_0}^{\top}(x_t-Jc)\|_2^2/\mu
=(x_t-Jc)^\top H(x_t-Jc).
\label{eq:pivot-weighted-error}
\end{equation}
We add ridge regularization $\lambda\|c\|_2^2$ with
$\lambda=0.003$, defined for coefficients in this
orthonormal basis.
Expanding Equation~\ref{eq:pivot-weighted-error},
differentiating the regularized objective with respect
to $c$, and setting the result to zero gives
\begin{align}
0
&=2J^\top H(Jc-x_t)+2\lambda c, \nonumber\\
(J^\top HJ+\lambda I_G)c
&=J^\top Hx_t, \nonumber\\
\coefficientvector[t,\lambda]
&=(J^\top HJ+\lambda I_G)^{-1}J^\top Hx_t.
\label{eq:pivot-ridge}
\end{align}
The exact regularized solve therefore depends on the
dense state metric $H$.
Constructing the required blocks of $H$ at every flush
is the expensive operation that we approximate below.

\paragraph{Step 2: Approximate the state metric.}
Choose a small configurable pivot count $P$ and set $p:=\min(G,P)$. Let
$Q\in\mathbb R^{V\times p}$ be an orthonormal basis for the first $p$
columns of the current sketch $\sketch={S'_0}^{\top}J$. The projector $QQ^\top$
keeps the part of the value space spanned by these pivot sketch vectors.
Define
\begin{equation}
L:=\frac{S'_0Q}{\sqrt\mu}\in\mathbb R^{K\times p}.
\label{eq:pivot-low-rank-factor}
\end{equation}
Its product is
\begin{equation}
LL^\top=S'_0QQ^\top{S'_0}^{\top}/\mu,
\label{eq:pivot-low-rank-metric}
\end{equation}
so $LL^\top$ retains the correlations in $H$ that pass through the selected
$p$-dimensional output subspace. A low-rank matrix alone can underestimate
the energy of individual state coordinates. We preserve the remaining
diagonal energy with
\begin{equation}
d_i:=\max\{H_{ii}-\|L_{i,:}\|_2^2,0\},
\qquad
\widehat H:=LL^\top+\operatorname{Diag}(d).
\label{eq:pivot-metric}
\end{equation}
Since $(LL^\top)_{ii}=\|L_{i,:}\|_2^2$, this choice makes
$\widehat H_{ii}=H_{ii}$ whenever the residual is nonnegative; the maximum
with zero protects against numerical roundoff. Thus, $\widehat H$ keeps a
small set of important cross-coordinate correlations and the per-coordinate
energy left outside that set.

\paragraph{Step 3: Substitute the approximation into the solve.}
Let
\begin{equation}
L_G:=J^\top L\in\mathbb R^{G\times p},
\qquad
D_0:=\operatorname{Diag}(J^\top d),
\qquad
D_G:=D_0+\lambda I_G.
\label{eq:pivot-small-factors}
\end{equation}
Substituting $\widehat H$ for $H$ in the two sides of the normal equation
gives
\begin{align}
J^\top\widehat HJ+\lambda I_G
&=L_GL_G^\top+D_G, \label{eq:pivot-approx-gram}\\
J^\top\widehat Hx_t
&=L_G(L^\top x_t)+J^\top\operatorname{Diag}(d)x_t.
\label{eq:pivot-approx-rhs}
\end{align}
These equalities follow by distributing $J^\top$ and $J$ over
$\widehat H=LL^\top+\operatorname{Diag}(d)$. They show explicitly that the
approximation is applied to both the Gram matrix and the query-dependent
right-hand side.

\paragraph{Step 4: Replace the $G\times G$ inverse with a $p\times p$ inverse.}
The approximate Gram matrix has the diagonal-plus-low-rank form
$D_G+L_GL_G^\top$. The Woodbury identity gives
\begin{equation}
(D_G+L_GL_G^\top)^{-1}
=D_G^{-1}-D_G^{-1}L_G
\left(I_p+L_G^\top D_G^{-1}L_G\right)^{-1}
L_G^\top D_G^{-1}.
\label{eq:pivot-woodbury}
\end{equation}
Only
\begin{equation}
T_P:=I_p+L_G^\top D_G^{-1}L_G\in\mathbb R^{p\times p}
\label{eq:pivot-small-system}
\end{equation}
requires a dense factorization. Because $p\leq P$ is small, this replaces a
dense $G\times G$ solve with diagonal operations in $G$ and a dense
$p\times p$ solve.

\paragraph{Step 5: Identify what is and is not approximated.}
The flush prepares the factors above in
$O(KVp+Vp^2+Gp^2+p^3)$ arithmetic, without forming the exact
$K\times G$ cross-product or factoring its $G\times G$ Gram matrix. The
recurrent state and the sketch remain exact; only the state metric used to
compute the coefficient vector is replaced by $\widehat H$.

\paragraph{Flush-step latency.}
Table~\ref{tab:coefficient-flush-latency} compares the exact regularized
coefficient map with the $P=4$ approximation, using the same sketch bases
and rank allocations. Both paths fuse coefficient construction into the
state update, without an additional full-state read.
By reducing the cross-product and Gram-system computation described above,
$P=4$ reduces complete flush-step latency by $3.92$--$4.51\times$
across Mamba-2, GDN, and KDA.

\begin{table}[htbp]
\centering
\small
\caption{Complete flush-step latency on one NVIDIA B300 at batch size 256
and $W=16$, with FP32 states and BF16 sketch/coefficient map.
Kernel measurements use synthetic activations; one CUDA graph holds the flush
step of all recurrent layers, and we report the median of five captures
divided by the layer count.
Exact solves the full regularized Gram system with the same $\lambda=0.003$ as $P=4$;
speedup is Exact latency divided by $P=4$ latency.}
\label{tab:coefficient-flush-latency}
\begin{tabular}{lrrrr}
\toprule
Model & $\bar G$ & Exact ($\mu$s) & $P=4$ ($\mu$s) & Speedup \\
\midrule
Nemotron 3 Super & 8 & 2,076.21 & 460.31 & $4.51\times$ \\
Qwen3.8 Flash-Next & 8 & 1,705.90 & 414.02 & $4.12\times$ \\
GLM 5.3 Flash & 8 & 2,650.73 & 675.79 & $3.92\times$ \\
\bottomrule
\end{tabular}
\end{table}

\clearpage
\section{SketchSSM Algorithm}
\label{app:sketchssm-algorithm}

Algorithm~\ref{alg:sketchssm} summarizes SketchSSM in
Section~\ref{sec:sketch} for one state head. The sketching matrix
$\sketchingmatrix$ is calibrated offline and fixed throughout inference.
After initialization, each window contains $W-1$ non-flush steps that
read $\sketch$ and $\coefficientmap$, followed by one flush step that reads
$\state$ and applies the buffered updates from $\mathcal B$ exactly.
The flush step computes the output from the updated full state and
constructs $\sketch,\coefficientmap$ before writing the state back to HBM.

\begin{algorithm}[H]
\caption{SketchSSM calibration and inference.}
\label{alg:sketchssm}
\begin{algorithmic}[1]
\Require Calibration distribution $\mathcal D$, mean rank budget $\bar G$, window size $W$
\Statex \textbf{Offline}
\State Calibrate $\sketchingmatrix^\star$ using Equation~\ref{eq:m-subspace-objective}
\State Allocate $G_{l,h}$ to each state head under $\bar G$ using Equation~\ref{eq:m-allocation}
\State Fix $\sketchingmatrix$ to the first $G_{l,h}$ columns of $\sketchingmatrix^\star$ for each head
\Statex \textbf{Inference}
\State Initialize the full state $\state$ from prefill
\State Construct the initial $\sketch,\coefficientmap$ from $\state$; initialize buffer $\mathcal B\gets\varnothing$
\While{decoding continues}
    \Statex \hspace{\algorithmicindent}\textbf{Non-flush steps: $W-1$ times per window}
    \For{$t=1,\ldots,W-1$}
        \State Append the current update inputs to $\mathcal B$
        \State Compute $\effectivequery$ and $\bufferoutput$ exactly from $\mathcal B$ using Equation~\ref{eq:m-reads}
        \State Read $\sketch,\coefficientmap$ from HBM
        \State $o_t\gets\sketch\coefficientmap\effectivequery+\bufferoutput$
        \State Emit $o_t$; terminate if decoding is complete
    \EndFor
    \Statex \hspace{\algorithmicindent}\textbf{Flush step: once per window, $t=W$}
    \State Append the current update inputs to $\mathcal B$
    \State Read the full state $\state$ from HBM
    \State Update $\state$ exactly using all buffered inputs from $\mathcal B$
    \State $o_t\gets\state^\top q_t$
    \State Construct the sketch $\sketch\gets\state^\top\sketchingmatrix$
    \State Construct the coefficient map $\coefficientmap$ from $\state$ via Equation~\ref{eq:m-coeff} and Appendix~\ref{app:pivot-coeff}
    \State Write $\state,\sketch,\coefficientmap$ to HBM
    \State Clear $\mathcal B$; emit $o_t$
\EndWhile
\end{algorithmic}
\end{algorithm}

\clearpage
\section{WY-Based Effective-Query and Coefficient Computation}
\label{app:projected-wy}

We adapt the WY representation used in chunkwise delta-rule
computation~\citep{yang2024parallelizing} to the decay-and-erase
transitions in Equation~\ref{eq:transition-update}.
We first derive an exact effective-query evaluation without a sketch or
coefficient map, then describe its implementation in coefficient space.

\subsection{Parallel Effective-Query Evaluation}
\label{app:wy-effective-query}
Within a window indexed by \(t=1,\ldots,W\), the effective query is
\begin{equation}
\effectivequery=M_{1:t}^\top q_t,
\qquad M_{1:t}=M_t\cdots M_1.
\label{eq:app-effective-query-definition}
\end{equation}
Applying each transposed transition to the query separately creates a
serial dependency chain. A WY-style factorization replaces this chain
with a sum of contributions from buffered tokens.

\paragraph{Factor the transition product.}
For GDN and KDA,
\begin{equation}
M_s=(I-\beta_s k_sk_s^\top)D_s
   =D_s-\beta_s k_s(k_s^\top D_s),
\label{eq:app-one-transition}
\end{equation}
where \(D_s\) is diagonal, including scalar decay as a special case.
Let \(d_{[t]}=\operatorname{diag}(D_t\cdots D_1)\) and
\begin{equation}
\ell_s(t):=(D_t\cdots D_{s+1})k_s,
\label{eq:app-left-factor}
\end{equation}
with an empty matrix product equal to the identity. The ordered product
has the exact form
\begin{equation}
M_{1:t}=\operatorname{Diag}(d_{[t]})
-\sum_{s=1}^{t}\ell_s(t)\pi_s^\top,
\label{eq:app-wy-form}
\end{equation}
where the erase factors satisfy
\begin{equation}
\pi_s=\beta_s\left(
d_{[s]}\odot k_s
-\sum_{j=1}^{s-1}\langle\ell_j(s),k_s\rangle\pi_j
\right)\in\mathbb R^K.
\label{eq:wy}
\end{equation}
To verify this form, the base case is
\(\pi_1=\beta_1(d_{[1]}\odot k_1)\).
At step \(s\), multiplication by \(D_s\) updates each existing left
factor to \(\ell_j(s)\). Multiplication by
\(I-\beta_s k_sk_s^\top\) then adds \(-k_s\pi_s^\top\), with
\(\pi_s\) given above. Thus the recurrence retains all cross terms
between erase operations.

\paragraph{Apply the product to the query.}
Transposing Equation~\ref{eq:app-wy-form} gives
\begin{equation}
\effectivequery=d_{[t]}\odot q_t
-\sum_{s=1}^{t}\pi_s\langle\ell_s(t),q_t\rangle.
\label{eq:app-effective-query-wy}
\end{equation}
Once the factors have been maintained as tokens arrive, their
contributions to the current query can be accumulated in parallel over
buffer slots. This evaluation is exact and requires no read of
\(\state\); it does not remove the causal dependence between arriving
tokens. For Mamba-2, the erase terms are absent and the effective query
reduces to \(d_{[t]}\odot q_t\).

\subsection{Coefficient Computation with Projected Factors}
\label{app:wy-coefficients}
We need \(\coefficientmap\effectivequery\), not
\(\effectivequery\) itself. Applying \(\coefficientmap\) to
Equation~\ref{eq:app-effective-query-wy} and defining
\(f_s:=\coefficientmap\pi_s\in\mathbb R^G\) gives
\begin{equation}
\coefficientvector^\star
=\coefficientmap(d_{[t]}\odot q_t)
-\sum_{s=1}^{t}f_s\langle\ell_s(t),q_t\rangle.
\label{eq:app-projected-query}
\end{equation}
Applying \(\coefficientmap\) to the recurrence for \(\pi_s\) likewise gives
\begin{equation}
f_s=\beta_s\left[
\coefficientmap(d_{[s]}\odot k_s)
-\sum_{j=1}^{s-1}
f_j\langle\ell_j(s),k_s\rangle
\right].
\label{eq:wyread}
\end{equation}
For a fixed coefficient map, these identities compute
\(\coefficientmap\effectivequery\) exactly; projecting the factors introduces
no additional approximation. Thus, once slot \(s\) is appended, later steps need only its
\(G\)-dimensional \(f_s\), rather than its \(K\)-dimensional \(\pi_s\).

\paragraph{Reusing buffered keys.}
Because the \(D_s\) matrices are diagonal,
\(\ell_s(t)=(d_{[t]}\oslash d_{[s]})\odot k_s\). Hence, for any
\(y\in\mathbb R^K\),
\begin{equation}
\langle\ell_s(t),y\rangle
=\langle k_s\oslash d_{[s]},d_{[t]}\odot y\rangle.
\label{eq:app-left-factor-dot}
\end{equation}
The kernel can compute the required scalar from the stored ring key and the
cumulative diagonal, so no additional \(K\)-vector is needed per slot. Both
sums in Equations~\ref{eq:app-projected-query} and~\ref{eq:wyread} are
parallel reductions over ring slots. In addition to the buffered inputs, this calculation uses the
coefficient map \(\coefficientmap\in\mathbb R^{G\times K}\) plus one
\(f_s\in\mathbb R^G\) per slot for erasing transitions. Mamba-2 has no
erase, so the \(f_s\) vectors and the two sums disappear. None of these
transformations changes the raw ring transitions used for the exact full-state
update at a flush.

\clearpage
\section{State Memory Traffic Reduction}
\label{sec:sketch-traffic}

Table~\ref{tab:m-traffic} compares amortized state-memory traffic,
including state-read and state-write accesses, in bytes per state head
per decode step. Standard reads and writes the full FP32 $K\times V$
state at every step, giving a reference cost of $8KV$ bytes.
Let $D:=4KV$ denote the full-state size, and let $C_{\mathrm{rate}}$
be the effective state compression ratio, including auxiliary metadata
such as quantization scales.

\para{State Write}
Standard writes $D$ bytes per step. Quantization and pruning reduce
this cost to $D/C_{\mathrm{rate}}$, as they update the compressed state
at every step. ReplaySSM and SketchSSM instead buffer updates and write
the full FP32 state once every $W$ steps. Their state-write traffic is
zero at non-flush steps and $D$ bytes at flush steps, averaging $D/W$
bytes per step.

\para{State Read}
Standard and ReplaySSM read $D$ bytes per step, while quantization and
pruning read $D/C_{\mathrm{rate}}$ bytes. SketchSSM stores the sketch
$\sketch$ and coefficient map $\coefficientmap$ in BF16, so reading
both at a non-flush step costs $2G(V+K)$ bytes. For GDN and KDA,
the BF16 projected erase vectors
(Appendix~\ref{app:wy-coefficients}) add $2GW$ bytes.
The total non-flush read cost is therefore
$\mathcal T_G:=2G(V+K+eW)$, with $e=0$ for Mamba-2 and $e=1$
for GDN and KDA. Each flush reads the full FP32 state to apply the
buffered updates and refresh these representations. Thus, SketchSSM's
average read cost is $B_G:=((W-1)\mathcal T_G+D)/W$.

\begin{table}[H]
\centering
\small
\setlength{\tabcolsep}{3pt}
\renewcommand{\arraystretch}{1.2}
\caption{Average state-memory traffic per state head per decode step,
including reads and writes, and reduction relative to Standard.
$D=4KV$, $B_G=((W-1)\mathcal T_G+D)/W$, and
$\mathcal T_G=2G(V+K+eW)$. Full states use FP32; sketches,
coefficient maps, projected erase vectors, and the state copies that
dense-fallback heads read at non-flush steps use BF16.}
\label{tab:m-traffic}
\begin{tabular*}{\textwidth}{@{\extracolsep{\fill}}lcccc@{}}
\toprule
Method & Read & Write & Total & Reduction \\
\midrule
Standard & $D$ & $D$ & $2D$ & $1$ \\
ReplaySSM & $D$ & $D/W$ & $D+D/W$ & $\frac{2W}{W+1}$ \\
Quant./Pruning & $D/C_{\mathrm{rate}}$ & $D/C_{\mathrm{rate}}$
& $2D/C_{\mathrm{rate}}$ & $C_{\mathrm{rate}}$ \\
SketchSSM & $B_G$ & $D/W$ & $B_G+D/W$
& $\frac{2D}{B_G+D/W}$ \\
\bottomrule
\end{tabular*}
\end{table}

For a sketched head, the state-memory traffic reduction is
\begin{equation}
R_G=\frac{8KV}{(W-1)\mathcal T_G/W+8KV/W}.
\end{equation}
For head-dependent ranks, we sum the byte costs across heads before
taking the ratio. A dense-fallback head keeps a BF16 copy of its state,
refreshed at each flush, and reads this copy ($D/2$ bytes) at non-flush
steps instead of a sketch; its average read cost is $((W-1)D/2+D)/W$,
and it writes $D/W$ bytes on average like other heads. At $W=16$, ReplaySSM reduces this traffic
by $32/17\approx1.88\times$. Quantization and pruning retain their
compression ratios because both reads and writes shrink proportionally.
This metric excludes the much smaller ring-buffer traffic and the writes of the sketch and coefficient map; these costs are reflected in the measured speedups.

\clearpage
\section{Offline Sketch-Rank Allocation}
\label{app:alloc-problem}

For each state head, we use the first $G$ columns of the offline basis
from Section~\ref{sec:sketch-def} to form a rank-$G$ sketch.
We keep the basis fixed and compare candidate ranks by their
state-read reconstruction errors and the loss sensitivity of those errors.

\paragraph{Head-level calibration.}
For head $j=(l,h)$, let $G_{\max}$ be the largest candidate rank and
$\sketch_{j,G_{\max}}\in\mathbb{R}^{V\times G_{\max}}$ its state sketch in
one calibration window. A thin QR factorization
$\sketch_{j,G_{\max}}=Q_jR_j$ evaluates all prefix residuals without solving
a separate least-squares problem for every rank:
\begin{equation}
\left\|\delta^o_{t,j}(G)\right\|_2^2
=\left\|\stateoutput[t,j]\right\|_2^2
-\left\|Q_{j,1:G}^{\top}\stateoutput[t,j]\right\|_2^2.
\label{eq:alloc-qr}
\end{equation}
On the same token, we record
$g_{t,j}=\nabla_{\stateoutput[t,j]}\ell$ and accumulate the
paired score $J_j(G)$ from Equation~\ref{eq:m-head-objective}. This
measures whether the discarded output is aligned with a loss-sensitive
direction, rather than multiplying independently averaged error and
gradient statistics. We also retain
$\varepsilon_j(G)=\mathbb E_t[\|\delta^o_{t,j}(G)\|_2^2]$ for the
gradient-free ablation. Basis fitting and allocation use disjoint
calibration examples.

The score is a separable first-order surrogate: it preserves the
sample-wise gradient--residual interaction within each head but omits
cross-head terms in
$(\sum_j g_{t,j}^{\top}\delta^o_{t,j})^2$. Its advantage is that the
resulting per-head curves can be optimized offline under an exact traffic
budget.

\paragraph{Budgeted optimization.}
Let $\tau_{\mathrm{sk}}:=2(V+K+eW)$ be the read traffic in bytes
contributed by one unit of sketch rank with BF16 storage.
A dense-fallback head instead reads the BF16 copy of its state at
non-flush steps, which costs $2KV$ bytes.
The largest admissible rank that is no more expensive than this dense read is
\[
G^{\star}:=
\min\left\{K,
\left\lfloor\frac{2KV}{\tau_{\mathrm{sk}}}\right\rfloor
\right\}.
\]
Each head chooses either a prefix rank in
$\{1,\ldots,G^{\star}\}$ or the dense fallback:
\begin{equation}
\begin{gathered}
\mathcal A:=\{1,\ldots,G^{\star},\mathrm{full}\},
\qquad
\tau(G):=G\tau_{\mathrm{sk}},
\qquad
\tau(\mathrm{full}):=2KV,
\qquad
J_j(\mathrm{full}):=0,\\
\min_{\{a_j\in\mathcal A\}}\sum_jJ_j(a_j)
\quad\text{s.t.}\quad
\sum_j\tau(a_j)\leq B,
\qquad
B:=N\bar G\tau_{\mathrm{sk}},
\end{gathered}
\label{eq:m-allocation}
\end{equation}
where $N$ is the number of state heads and $\bar G$ is the mean
sketch-rank budget. The flush read cost is identical for every action
and therefore does not affect this optimization.

The measured rank curves need not have diminishing returns, so a
sequential greedy rule is not guaranteed to minimize
Equation~\ref{eq:m-allocation}. We instead introduce a traffic price
$\mu\geq0$. For any price,
\begin{equation}
D(\mu):=
\sum_j\min_{a\in\mathcal A}\{J_j(a)+\mu\tau(a)\}-\mu B
\label{eq:alloc-dual}
\end{equation}
is a lower bound on the budgeted optimum. The price makes the choices
independent across heads: each head selects its lowest penalized score,
and bisection adjusts $\mu$ until their total traffic reaches the budget.
We retain the best feasible table encountered and spend any remaining
discrete slack through improving single-head substitutions.

If the returned feasible table has objective $P$, then
$(P-D(\mu))/P$ certifies its relative gap from the optimum of the
measured separable objective. This optimization runs only during offline
calibration; inference stores and applies the resulting fixed rank table.

\begin{figure}[htbp]
\centering
\includegraphics[width=\textwidth]{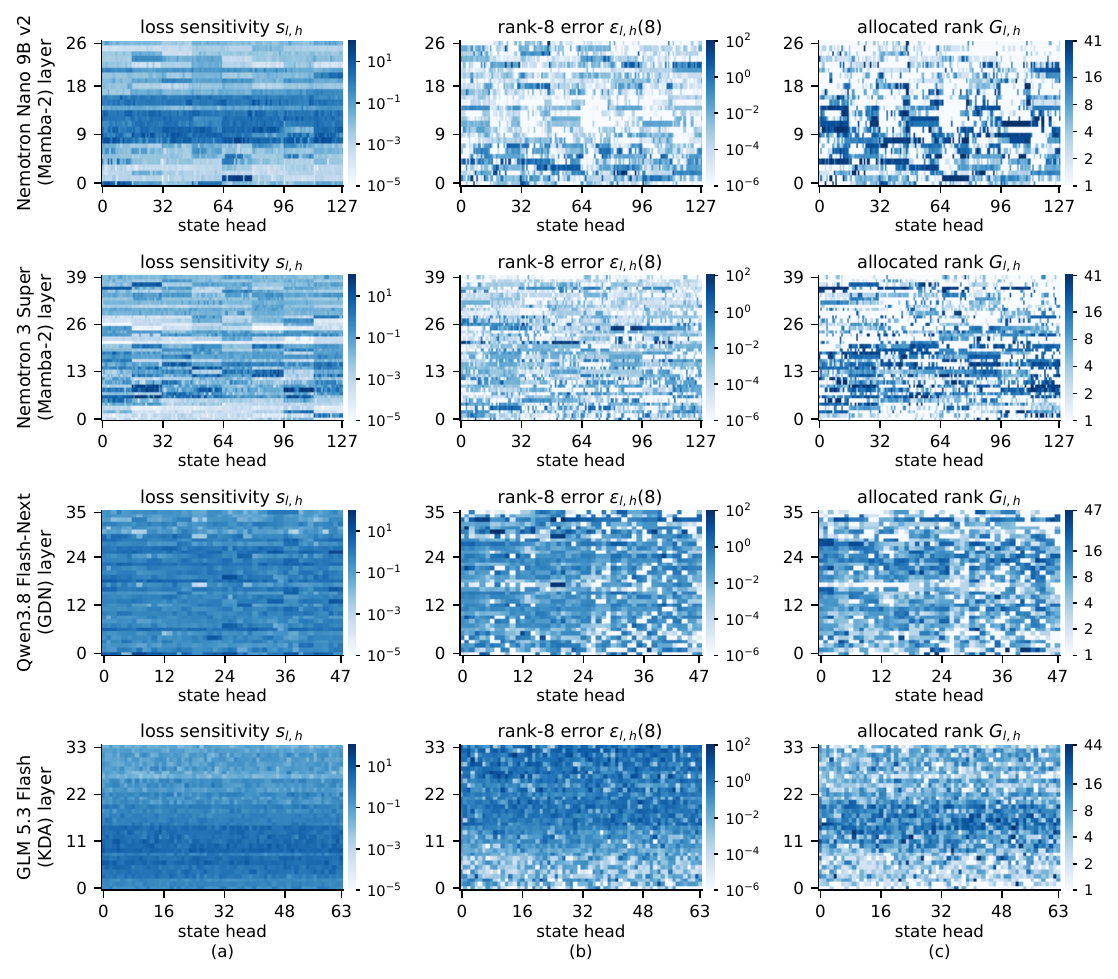}
\caption{Head-level rank allocation at $\bar G=8$ for each model (rows).
(a) Loss sensitivity $s_{l,h}=\mathbb E_t[\|g_{t,(l,h)}\|_2^2]$ of head $h$ in layer $l$.
(b) Rank-8 state-read error $\varepsilon_{l,h}(8)$.
(c) Allocated rank $G_{l,h}$; heads that retain the dense state are shown at the
maximum color intensity.
Nemotron Nano: ranks 1--41, 108 dense heads of 3,456 ($9.94\times$ traffic reduction);
Nemotron Super: 1--41, 159 of 5,120 ($9.39\times$);
Qwen3.8 Flash-Next: 1--47, 1 of 1,728 ($10.68\times$);
GLM 5.3 Flash: 1--44, 3 of 2,176 ($10.68\times$).
Sensitivity and error are normalized by their mean over heads.}
\label{fig:sketch-alloc}
\end{figure}

\begin{figure}[htbp]
\centering
\includegraphics[width=\textwidth]{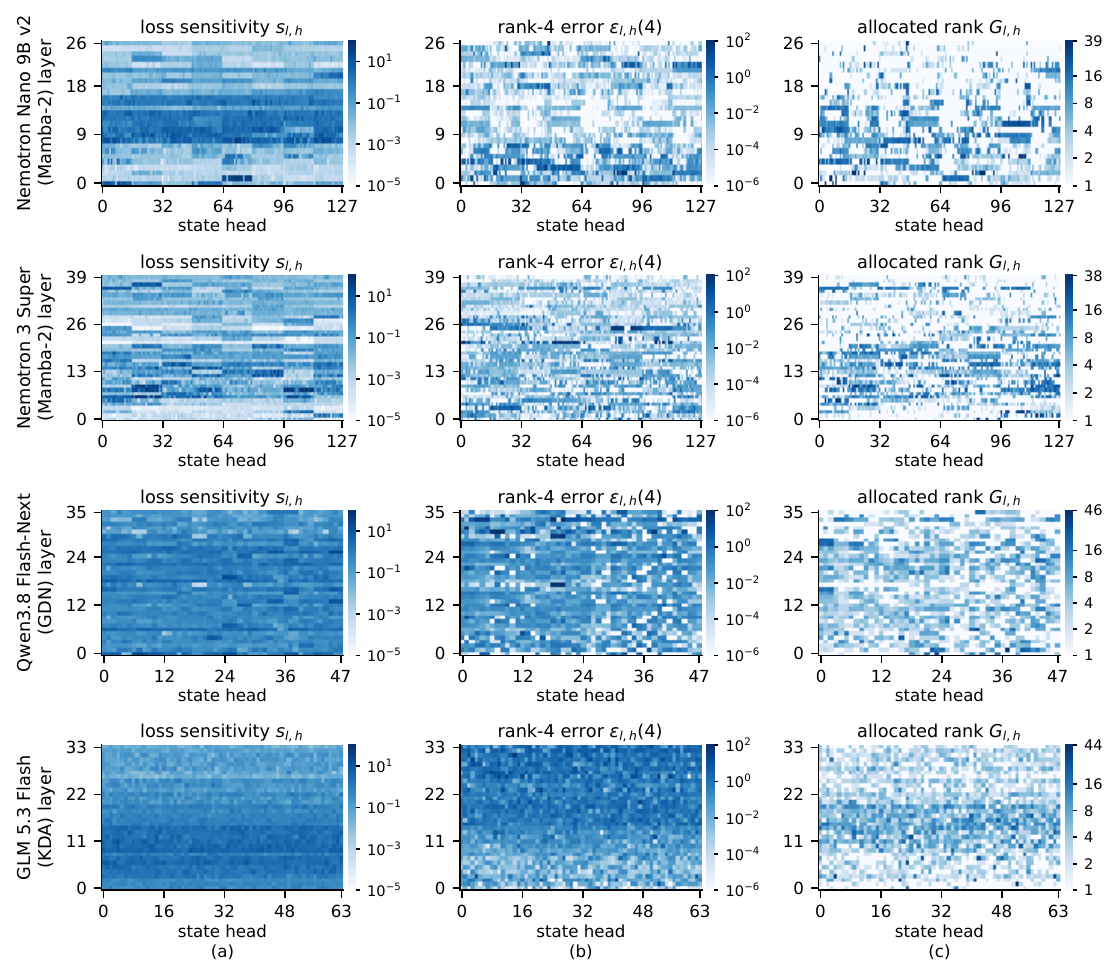}
\caption{Head-level rank allocation at $\bar G=4$, as in Fig.~\ref{fig:sketch-alloc}.
(b) shows the rank-4 state-read error $\varepsilon_{l,h}(4)$.
Nemotron Nano: ranks 1--39, 12 dense heads of 3,456 ($12.26\times$ traffic reduction);
Nemotron Super: 1--38, 5 of 5,120 ($11.84\times$);
Qwen3.8 Flash-Next: 1--46, no dense heads ($12.81\times$);
GLM 5.3 Flash: 1--44, no dense heads ($12.81\times$).}
\label{fig:sketch-alloc-g4}
\end{figure}

\paragraph{Allocated ranks.}
Figs.~\ref{fig:sketch-alloc} and~\ref{fig:sketch-alloc-g4} show the head-level
allocation for all four models at $\bar G=8$ and $\bar G=4$. Loss sensitivity and
state-read error vary widely across layers and heads, and the allocation assigns
larger ranks to heads where both are high.

\clearpage
\section{Evaluation Details}
\label{app:accuracy-results}

\subsection{Detailed Evaluation Setup}
\label{app:evaluation-setup}

\paragraph{Offline calibration.}
All calibration uses only general-domain text from WikiText-2
(\texttt{wikitext-2-raw-v1})~\citep{wiki2}, tokenized with each model's tokenizer.
This corpus is out of domain for all evaluated tasks (competition mathematics,
graduate-level science questions, code generation, and the RULER retrieval tasks),
and the same calibration is used for every benchmark without any benchmark data.
For basis calibration, we feed 520 prompts of 512 tokens from the
training split and let each model greedily generate 256 further tokens.
Over the generated tokens (excluding the first window), we collect
window-boundary states and effective queries with $W=16$ and estimate
the state-weighted query statistics described
in Appendix~\ref{app:offline-sketch-proof} to construct
the sketching matrix.
For rank allocation, we use 64 sequences of 256 tokens from the
validation split and score the 8,192 token positions that follow a
128-token warmup.
We compute gradients of the mean next-token cross-entropy loss
and pair each head's output gradient with its reconstruction
residual to score candidate ranks
(Appendix~\ref{app:alloc-problem}).
These scores use ideal least-squares reconstruction,
while inference uses the coefficient approximation in
Appendix~\ref{app:pivot-coeff}.
Calibration leaves model weights unchanged, and the resulting
sketching matrices and rank allocations remain fixed during inference.
We release the calibrated sketching matrices and the rank-allocation scores
of all four models on Hugging Face\footnote{\url{https://huggingface.co/SketchSSM}};
the rank table for a given mean budget $\bar G$ is computed from these scores
when the model is loaded.

\paragraph{Accuracy evaluation setup.}
SketchSSM settings are mean sketch-rank budgets $\bar G$.
The coefficient approximation uses $P=4$ pivots for all four models.
SketchSSM maintains the full recurrent state in FP32
and stores both the sketch $\sketch$ and the coefficient
map $\coefficientmap$ in BF16.
The coefficient solve uses a ridge of $\lambda=0.003$.
ReplaySSM and SketchSSM use a window size of $W=16$.
Standard, ReplaySSM, and SketchSSM are all evaluated on vLLM 0.30.0
with our SketchSSM integration, on one NVIDIA B300 GPU.
For the ReplaySSM baseline, we use the official implementation
integrated into vLLM for Mamba-2-based models, and for GDN-based models
we port the authors' ReplaySSM implementation\footnote{\url{https://github.com/Johnny-Liou/ReplaySSM}}
to vLLM 0.30.0. As no official implementation is available for KDA,
we implement the same window replay for the KDA-based model.
We report full-state results for both Standard execution
(without ReplaySSM) and ReplaySSM.
For Standard, ReplaySSM, and SketchSSM, we sample with each model's
recommended settings (temperature $0.6$ and top-$p$ $0.95$ for Nemotron Nano;
temperature $1.0$ and top-$p$ $0.95$ for the other models), using a fixed
sampling seed per request, and report pass@1 averaged over eight seeds
(0--7) for AIME25 and three seeds (0--2) for the other benchmarks.
The state-quantization and state-pruning baselines were evaluated earlier on
vLLM 0.27.0 (Mamba-2 models) and vLLM 0.28.1 (GDN and KDA models). We evaluate MATH-500 ($500$ problems), AIME25 ($30$), GPQA Diamond
($198$), and LiveCodeBench ($315$, split \texttt{v5\_2407\_2412}).
Generation caps are $32{,}768$ tokens for Nemotron Nano and $65{,}536$
for Nemotron 3 Super, Qwen3.8 Flash-Next, and GLM 5.3 Flash.

\paragraph{Speed evaluation setup.}
We measure kernel latency on one NVIDIA B300 with $W=16$
and batch sizes 128, 256, and 512
(Fig.~\ref{fig:super-kernel-speedup}).
The sketch-rank budgets are $\bar G\in\{8,4\}$ for all three models.
The measurement runs the recurrent kernels only, with every recurrent layer
(40, 36, and 34 for Super, Qwen, and GLM) holding its own state, buffers, and
calibrated sketch. One CUDA graph holds one decode step of all layers; we report
the median of five captures divided by the layer count. A non-flush step
launches no flush kernel.
The Standard Mamba-2 baseline uses a launch configuration of vLLM's
\texttt{selective\_state\_update} tuned for B300, as vLLM ships none for this GPU.
ReplaySSM uses vLLM's implementation for Mamba-2 and the authors' implementation ported to vLLM 0.30.0 for GDN.
Because no official ReplaySSM implementation is available for KDA,
we implement the KDA ReplaySSM baseline ourselves, running its window
bookkeeping once per step for all layers.
Flush overhead is measured against the same SketchSSM flush kernel
with sketch and coefficient-map construction disabled.
For Nemotron Super, we also measure decode throughput at 2K and 8K
input lengths, up to each method's maximum batch size
(Fig.~\ref{fig:super-e2e-speedup}).
Each run releases a cohort of $B$ prompts of $C$ tokens into decode together
and runs 128 decode steps; we report the mean model-forward GPU time over
steps 16--111, averaged over two runs, with every method on vLLM's Model
Runner V2 and one shared FlashInfer autotuning result.
The latency breakdown comes from one profiled cohort at batch size 512.

\subsection{Accuracy Results on Decode Benchmarks}
\label{app:decode-results}

Tables~\ref{tab:accuracy-nano}--\ref{tab:accuracy-glm} report the
data plotted in Fig.~\ref{fig:pareto}. Accuracy is in percent, and
verbosity is the mean number of generated tokens normalized by each model's
baseline. The traffic column
includes flush traffic and DSQ's dynamic scale-factor loads.

\para{Verbosity}
We measure normalized verbosity as the mean generated-token count
relative to Standard for each benchmark, averaged across the four
benchmarks.
At $\bar G=8$ (9.4--10.7$\times$ traffic reduction), normalized verbosity is
$1.04$, $0.95$, $1.00$, and $0.86$ for Nemotron Nano, Nemotron Super,
Qwen3.8 Flash-Next, and GLM 5.3 Flash, respectively.
It is $1.11$, $0.98$, $1.15$, and $0.87$ at $\bar G=4$
(11.8--12.8$\times$),
and rises to $1.21$, $1.03$, $1.58$, and $1.06$ at $\bar G=2$
(13.6--14.2$\times$).
These results illustrate a tradeoff between state-access traffic
reduction and generated response length.

\begin{table}[H]
\centering
\scriptsize
\caption{Nemotron Nano 9B v2: numerical results underlying Fig.~\ref{fig:pareto}. ReplaySSM is additionally reported. Accuracies are in percent; token counts are mean generated tokens. Standard, ReplaySSM, and SketchSSM report pass@1 averaged over eight seeds for AIME25 and three seeds for the other benchmarks; subscripts are the standard deviation across seeds (for Average, the per-benchmark deviations combined as $\sqrt{\sum_b\sigma_b^2}/4$). GHOST and DSQ are single runs. Traffic is state-read plus state-write traffic reduction relative to Standard. Average is the unweighted mean accuracy across the four benchmarks.} 
\label{tab:accuracy-nano}
\setlength{\tabcolsep}{1pt}
\begin{tabular*}{\textwidth}{@{\extracolsep{\fill}}llr*{4}{rr}r@{}}
\toprule
& & & \multicolumn{2}{c}{MATH-500} & \multicolumn{2}{c}{AIME25}
& \multicolumn{2}{c}{GPQA-D} & \multicolumn{2}{c}{LiveCodeBench} & Average \\
\cmidrule(lr){4-5}\cmidrule(lr){6-7}\cmidrule(lr){8-9}\cmidrule(lr){10-11}
Method & Setting & Traffic & Acc. & Tokens & Acc. & Tokens & Acc. & Tokens & Acc. & Tokens & Acc. \\
\midrule
\textbf{Standard} & \textbf{full-state} & {\boldmath$1.00\times$} & $97.33_{\pm 0.42}$ & 5,257 & $71.67_{\pm 3.56}$ & 18,868 & $59.43_{\pm 1.17}$ & 13,499 & $68.99_{\pm 1.20}$ & 14,073 & {\boldmath$74.36_{\pm 0.99}$} \\
ReplaySSM & full-state & $1.88\times$ & $97.33_{\pm 0.61}$ & 5,142 & $73.75_{\pm 5.18}$ & 18,344 & $60.10_{\pm 1.01}$ & 13,288 & $69.42_{\pm 2.40}$ & 14,174 & $75.15_{\pm 1.46}$ \\
\midrule
\textbf{SketchSSM} & {\boldmath$\bar G=8$} & {\boldmath$9.94\times$} & $97.07_{\pm 0.58}$ & 5,548 & $71.67_{\pm 3.09}$ & 19,341 & $60.77_{\pm 0.77}$ & 13,969 & $68.15_{\pm 1.63}$ & 14,915 & {\boldmath$74.41_{\pm 0.91}$} \\
SketchSSM & $\bar G=4$ & $12.26\times$ & $96.73_{\pm 0.31}$ & 5,801 & $70.42_{\pm 5.18}$ & 20,336 & $58.25_{\pm 0.58}$ & 14,257 & $62.33_{\pm 1.50}$ & 16,705 & $71.93_{\pm 1.36}$ \\
SketchSSM & $\bar G=2$ & $13.88\times$ & $94.87_{\pm 0.64}$ & 6,510 & $60.83_{\pm 4.27}$ & 21,692 & $59.60_{\pm 1.01}$ & 14,673 & $57.67_{\pm 1.81}$ & 18,963 & $68.24_{\pm 1.20}$ \\
\midrule
\textbf{GHOST} & \textbf{50\%} & {\boldmath$2.00\times$} & 66.00 & 11,156 & 12.50 & 26,059 & 22.70 & 18,283 & 23.50 & 24,053 & \textbf{31.18} \\
GHOST & 62.5\% & $2.67\times$ & 51.00 & 14,995 & 2.50 & 30,924 & 13.60 & 22,710 & 14.60 & 26,609 & 20.43 \\
GHOST & 75\% & $4.00\times$ & 1.00 & 27,447 & 0.00 & 28,530 & 0.00 & 25,613 & 0.00 & 31,098 & 0.25 \\
\midrule
\textbf{DSQ} & \textbf{8-bit} & {\boldmath$3.84\times$} & 85.40 & 9,585 & 17.50 & 30,167 & 40.90 & 21,377 & 31.40 & 24,748 & \textbf{43.80} \\
DSQ & 6-bit & $5.06\times$ & 70.60 & 15,208 & 8.30 & 31,599 & 27.80 & 24,186 & 12.70 & 28,807 & 29.85 \\
DSQ & 4-bit & $7.40\times$ & 9.40 & 29,462 & 0.00 & 32,102 & 0.00 & 32,185 & 5.40 & 30,798 & 3.70 \\
\bottomrule
\end{tabular*}
\end{table}

\begin{table}[H]
\centering
\scriptsize
\caption{Nemotron 3 Super 120B-A12B: numerical results underlying Fig.~\ref{fig:pareto}. ReplaySSM is additionally reported.
Accuracies are in percent; token counts are mean generated tokens. Standard, ReplaySSM, and SketchSSM report pass@1 averaged over eight seeds for AIME25 and three seeds for the other benchmarks; subscripts are the standard deviation across seeds (for Average, the per-benchmark deviations combined as $\sqrt{\sum_b\sigma_b^2}/4$). GHOST and DSQ are single runs.
Traffic is state-read plus state-write traffic reduction relative to Standard.
Average is the unweighted mean accuracy across the four benchmarks.}
\label{tab:accuracy-super}
\setlength{\tabcolsep}{1pt}
\begin{tabular*}{\textwidth}{@{\extracolsep{\fill}}llr*{4}{rr}r@{}}
\toprule
& & & \multicolumn{2}{c}{MATH-500} & \multicolumn{2}{c}{AIME25}
& \multicolumn{2}{c}{GPQA-D} & \multicolumn{2}{c}{LiveCodeBench} & Average \\
\cmidrule(lr){4-5}\cmidrule(lr){6-7}\cmidrule(lr){8-9}\cmidrule(lr){10-11}
Method & Setting & Traffic & Acc. & Tokens & Acc. & Tokens & Acc. & Tokens & Acc. & Tokens & Acc. \\
\midrule
\textbf{Standard} & \textbf{full-state} & {\boldmath$1.00\times$} & $98.27_{\pm 0.46}$ & 3,439 & $82.50_{\pm 2.95}$ & 22,635 & $77.27_{\pm 0.87}$ & 18,868 & $80.21_{\pm 0.80}$ & 19,445 & {\boldmath$84.56_{\pm 0.80}$} \\
ReplaySSM & full-state & $1.88\times$ & $98.27_{\pm 0.23}$ & 3,309 & $84.17_{\pm 4.96}$ & 23,683 & $76.09_{\pm 1.05}$ & 18,650 & $79.47_{\pm 2.11}$ & 19,572 & $84.50_{\pm 1.37}$ \\
\midrule
\textbf{SketchSSM} & {\boldmath$\bar G=8$} & {\boldmath$9.39\times$} & $98.40_{\pm 0.35}$ & 3,148 & $82.08_{\pm 3.96}$ & 23,456 & $77.44_{\pm 2.49}$ & 17,142 & $80.74_{\pm 0.80}$ & 18,249 & {\boldmath$84.67_{\pm 1.19}$} \\
SketchSSM & $\bar G=4$ & $11.84\times$ & $98.20_{\pm 0.53}$ & 3,512 & $85.00_{\pm 2.52}$ & 23,134 & $78.45_{\pm 0.29}$ & 17,578 & $76.08_{\pm 2.25}$ & 18,634 & $84.43_{\pm 0.86}$ \\
SketchSSM & $\bar G=2$ & $13.61\times$ & $97.87_{\pm 0.42}$ & 4,111 & $80.83_{\pm 2.36}$ & 25,879 & $75.42_{\pm 0.77}$ & 15,677 & $72.91_{\pm 1.75}$ & 18,807 & $81.76_{\pm 0.77}$ \\
\midrule
\textbf{GHOST} & \textbf{50\%} & {\boldmath$2.00\times$} & 93.80 & 7,904 & 49.60 & 43,232 & 60.10 & 17,834 & 54.30 & 23,895 & \textbf{64.45} \\
GHOST & 62.5\% & $2.67\times$ & 72.20 & 21,121 & 11.70 & 60,528 & 48.00 & 26,047 & 26.30 & 31,228 & 39.55 \\
GHOST & 75\% & $4.00\times$ & 26.40 & 45,937 & 0.00 & 63,443 & 5.60 & 56,059 & 3.20 & 59,983 & 8.80 \\
\midrule
\textbf{DSQ} & \textbf{8-bit} & {\boldmath$3.82\times$} & 96.60 & 5,075 & 62.50 & 33,506 & 67.20 & 22,668 & 69.80 & 24,807 & \textbf{74.03} \\
DSQ & 6-bit & $5.02\times$ & 93.40 & 7,431 & 52.50 & 39,523 & 62.60 & 25,577 & 57.10 & 32,716 & 66.40 \\
DSQ & 4-bit & $7.31\times$ & 68.80 & 21,032 & 5.80 & 60,756 & 33.30 & 42,752 & 17.80 & 51,194 & 31.42 \\
\bottomrule
\end{tabular*}
\end{table}

\begin{table}[H]
\centering
\scriptsize
\caption{Qwen3.8 Flash-Next: numerical results underlying Fig.~\ref{fig:pareto}. ReplaySSM is additionally reported.
Accuracies are in percent; token counts are mean generated tokens. Standard, ReplaySSM, and SketchSSM report pass@1 averaged over eight seeds for AIME25 and three seeds for the other benchmarks; subscripts are the standard deviation across seeds (for Average, the per-benchmark deviations combined as $\sqrt{\sum_b\sigma_b^2}/4$). DRRQR and DSQ are single runs.
Traffic is state-read plus state-write traffic reduction relative to Standard.
Average is the unweighted mean accuracy across the four benchmarks.}
\label{tab:accuracy-qwen}
\setlength{\tabcolsep}{1pt}
\begin{tabular*}{\textwidth}{@{\extracolsep{\fill}}llr*{4}{rr}r@{}}
\toprule
& & & \multicolumn{2}{c}{MATH-500} & \multicolumn{2}{c}{AIME25}
& \multicolumn{2}{c}{GPQA-D} & \multicolumn{2}{c}{LiveCodeBench} & Average \\
\cmidrule(lr){4-5}\cmidrule(lr){6-7}\cmidrule(lr){8-9}\cmidrule(lr){10-11}
Method & Setting & Traffic & Acc. & Tokens & Acc. & Tokens & Acc. & Tokens & Acc. & Tokens & Acc. \\
\midrule
\textbf{Standard} & \textbf{full-state} & {\boldmath$1.00\times$} & $99.07_{\pm 0.23}$ & 1,434 & $94.17_{\pm 1.54}$ & 19,455 & $87.88_{\pm 1.82}$ & 14,736 & $81.48_{\pm 0.92}$ & 22,039 & {\boldmath$90.65_{\pm 0.64}$} \\
ReplaySSM & full-state & $1.88\times$ & $99.07_{\pm 0.12}$ & 1,456 & $94.17_{\pm 2.36}$ & 19,714 & $91.25_{\pm 0.77}$ & 14,921 & $79.58_{\pm 2.16}$ & 21,819 & $91.01_{\pm 0.82}$ \\
\midrule
\textbf{SketchSSM} & {\boldmath$\bar G=8$} & {\boldmath$10.68\times$} & $99.00_{\pm 0.20}$ & 1,474 & $94.58_{\pm 2.48}$ & 19,560 & $89.90_{\pm 0.87}$ & 14,232 & $79.05_{\pm 1.38}$ & 21,665 & {\boldmath$90.63_{\pm 0.74}$} \\
SketchSSM & $\bar G=4$ & $12.81\times$ & $99.07_{\pm 0.31}$ & 1,786 & $89.17_{\pm 2.95}$ & 22,457 & $88.89_{\pm 1.52}$ & 15,947 & $77.25_{\pm 0.18}$ & 25,061 & $88.59_{\pm 0.83}$ \\
SketchSSM & $\bar G=2$ & $14.23\times$ & $98.47_{\pm 0.58}$ & 2,819 & $85.00_{\pm 5.63}$ & 29,609 & $82.66_{\pm 1.05}$ & 21,334 & $66.46_{\pm 0.97}$ & 30,263 & $83.15_{\pm 1.46}$ \\
\midrule
\textbf{DRRQR} & \textbf{50\%} & {\boldmath$2.00\times$} & 95.00 & 2,008 & 89.58 & 20,705 & 79.29 & 11,735 & 66.98 & 24,840 & \textbf{82.72} \\
DRRQR & 62.5\% & $2.67\times$ & 92.20 & 3,276 & 78.75 & 22,805 & 74.24 & 10,918 & 51.11 & 28,072 & 74.08 \\
DRRQR & 75\% & $4.00\times$ & 9.40 & 1,548 & 0.42 & 5,926 & 2.53 & 1,879 & 0.00 & 1,969 & 3.09 \\
\midrule
\textbf{DSQ} & \textbf{8-bit} & {\boldmath$3.88\times$} & 98.20 & 1,900 & 71.25 & 29,344 & 75.25 & 21,650 & 68.25 & 26,888 & \textbf{78.24} \\
DSQ & 6-bit & $5.12\times$ & 94.80 & 2,952 & 38.33 & 24,680 & 48.99 & 17,251 & 37.78 & 19,475 & 54.98 \\
DSQ & 4-bit & $7.53\times$ & 57.40 & 2,162 & 0.42 & 4,872 & 9.09 & 3,928 & 13.97 & 3,756 & 20.22 \\
\bottomrule
\end{tabular*}
\end{table}

\begin{table}[H]
\centering
\scriptsize
\caption{GLM 5.3 Flash: numerical results underlying Fig.~\ref{fig:pareto}. ReplaySSM is additionally reported.
Accuracies are in percent; token counts are mean generated tokens. Standard, ReplaySSM, and SketchSSM report pass@1 averaged over eight seeds for AIME25 and three seeds for the other benchmarks; subscripts are the standard deviation across seeds (for Average, the per-benchmark deviations combined as $\sqrt{\sum_b\sigma_b^2}/4$). GHOST and DSQ are single runs.
Traffic is state-read plus state-write traffic reduction relative to Standard.
Average is the unweighted mean accuracy across the four benchmarks.}
\label{tab:accuracy-glm}
\setlength{\tabcolsep}{1pt}
\begin{tabular*}{\textwidth}{@{\extracolsep{\fill}}llr*{4}{rr}r@{}}
\toprule
& & & \multicolumn{2}{c}{MATH-500} & \multicolumn{2}{c}{AIME25}
& \multicolumn{2}{c}{GPQA-D} & \multicolumn{2}{c}{LiveCodeBench} & Average \\
\cmidrule(lr){4-5}\cmidrule(lr){6-7}\cmidrule(lr){8-9}\cmidrule(lr){10-11}
Method & Setting & Traffic & Acc. & Tokens & Acc. & Tokens & Acc. & Tokens & Acc. & Tokens & Acc. \\
\midrule
\textbf{Standard} & \textbf{full-state} & {\boldmath$1.00\times$} & $97.73_{\pm 0.61}$ & 1,239 & $81.67_{\pm 4.36}$ & 22,038 & $90.24_{\pm 0.77}$ & 10,790 & $74.50_{\pm 2.64}$ & 24,524 & {\boldmath$86.03_{\pm 1.30}$} \\
ReplaySSM & full-state & $1.88\times$ & $97.40_{\pm 0.40}$ & 1,165 & $86.25_{\pm 5.47}$ & 18,556 & $89.56_{\pm 0.58}$ & 10,797 & $74.18_{\pm 0.73}$ & 25,005 & $86.85_{\pm 1.39}$ \\
\midrule
\textbf{SketchSSM} & {\boldmath$\bar G=8$} & {\boldmath$10.68\times$} & $97.60_{\pm 0.20}$ & 1,119 & $85.42_{\pm 4.34}$ & 18,984 & $88.89_{\pm 1.52}$ & 10,014 & $82.01_{\pm 1.28}$ & 18,647 & {\boldmath$88.48_{\pm 1.19}$} \\
SketchSSM & $\bar G=4$ & $12.81\times$ & $97.53_{\pm 0.64}$ & 1,170 & $86.67_{\pm 4.71}$ & 17,298 & $85.69_{\pm 1.05}$ & 10,720 & $79.58_{\pm 1.94}$ & 18,153 & $87.37_{\pm 1.31}$ \\
SketchSSM & $\bar G=2$ & $14.23\times$ & $95.40_{\pm 0.69}$ & 1,717 & $80.00_{\pm 6.90}$ & 19,880 & $81.14_{\pm 2.96}$ & 12,944 & $67.09_{\pm 1.63}$ & 18,740 & $80.91_{\pm 1.93}$ \\
\midrule
\textbf{GHOST} & \textbf{50\%} & {\boldmath$2.00\times$} & 66.60 & 16,894 & 4.58 & 58,929 & 25.25 & 40,181 & 9.52 & 59,656 & \textbf{26.49} \\
GHOST & 62.5\% & $2.67\times$ & 37.20 & 35,219 & 0.83 & 63,749 & 7.07 & 54,124 & 2.54 & 62,403 & 11.91 \\
GHOST & 75\% & $4.00\times$ & 3.20 & 62,161 & 0.00 & 64,948 & 3.03 & 61,297 & 0.00 & 64,747 & 1.56 \\
\midrule
\textbf{DSQ} & \textbf{8-bit} & {\boldmath$3.88\times$} & 96.20 & 998 & 82.50 & 20,791 & 88.38 & 10,625 & 70.79 & 24,455 & \textbf{84.47} \\
DSQ & 6-bit & $5.12\times$ & 93.60 & 2,744 & 37.50 & 29,547 & 55.05 & 21,322 & 36.51 & 30,678 & 55.66 \\
DSQ & 4-bit & $7.53\times$ & 57.80 & 15,839 & 0.00 & 36,383 & 7.07 & 45,971 & 0.32 & 48,541 & 16.30 \\
\bottomrule
\end{tabular*}
\end{table}

\FloatBarrier

\clearpage
\subsection{Recall Evaluation}
\label{app:recall-results}

\paragraph{Evaluation setup.}
We evaluate long-context recall on four retrieval tasks from
RULER~\citep{hsieh2024ruler}: Single UUID
(\texttt{niah\_single\_3}), Multi-key
(\texttt{niah\_multikey\_2}), Multi-value
(\texttt{niah\_multivalue}), and Multi-query
(\texttt{niah\_multiquery}).
These tasks cover retrieving a UUID, identifying a value among distractor
key--value pairs, retrieving four values for one key, and answering four
queries, respectively. We use 500 examples per task at a 16K context budget,
counted with each model's tokenizer, and reuse identical inputs across
methods within each model. Generation is greedy with a cap of 128 tokens.
We follow RULER's answer-coverage metric: the fraction of reference answers
found in the generated text by case-insensitive substring matching,
averaged over examples and reported as a percentage. Average is the
unweighted mean of the four task scores.
We wrap the task in each model's chat template and configure direct-answer
generation. We retain RULER's default answer prefix, except for
Qwen3.8 Flash-Next, for which we observed premature termination with this
prefix. We therefore omit it consistently across all evaluated methods
for this model, while retaining its native non-thinking template.
For GLM 5.3 Flash, we use a custom direct-answer template that removes the
reasoning-effort instruction and closes the initial thinking block.
We compare Standard execution, ReplaySSM with $W=16$,
SketchSSM with $W=16$ and $P=4$, and state pruning at 50\%, 62.5\%, and
75\% sparsity. SketchSSM uses fixed offline-calibrated bases and rank
allocations, without calibration on the recall examples; its full state
is FP32 and its sketch and coefficient map are stored in BF16.

\paragraph{Results.}
On Nemotron Nano, SketchSSM achieves average recall scores of
97.92--98.03\% at $\bar G\in\{8,4\}$, compared with 98.29\% for
Standard execution and 98.25\% for ReplaySSM.
The largest task-level decrease occurs on Multi-value at $\bar G=4$,
from 97.45\% for Standard execution to 94.90\%.
On Nemotron Super, SketchSSM matches both full-state methods at 100\%
on all four tasks for each reported rank.
Qwen3.8 Flash-Next also scores 100\% on all four tasks for
Standard execution, ReplaySSM, and SketchSSM at both reported ranks.
On GLM 5.3 Flash, SketchSSM achieves average scores of 99.70--99.74\%,
compared with 99.40\% for Standard execution and 99.46\% for ReplaySSM.
State pruning causes larger losses: at 75\% sparsity, average scores
are 76.70\% for Nemotron Nano, 53.03\% for Nemotron Super,
43.05\% for Qwen3.8 Flash-Next, and 46.78\% for GLM 5.3 Flash.
The task-level results below expose differences that an average alone
would obscure.

\begin{table}[H]
\centering
\footnotesize
\caption{Recall scores (\%) on four RULER retrieval tasks at 16K context for Nemotron Nano. Traffic is the state-read plus state-write traffic reduction relative to Standard, using the same accounting as the accuracy tables. Average is the unweighted mean of the four task scores.}
\label{tab:recall-nano}
\setlength{\tabcolsep}{3pt}
\begin{tabular*}{\textwidth}{@{\extracolsep{\fill}}llrrrrrr@{}}
\toprule
Method & Setting & Traffic & Single UUID & Multi-key & Multi-value & Multi-query & Average \\
\midrule
Standard & full-state & $1.00\times$ & 100.00 & 99.80 & 97.45 & 95.90 & 98.29 \\
ReplaySSM & full-state & $1.88\times$ & 100.00 & 99.80 & 97.20 & 96.00 & 98.25 \\
\midrule
SketchSSM & $\bar G=8$ & $9.94\times$ & 100.00 & 99.80 & 96.75 & 95.55 & 98.03 \\
SketchSSM & $\bar G=4$ & $12.26\times$ & 100.00 & 99.80 & 94.90 & 97.00 & 97.92 \\
\midrule
GHOST & 50\% & $2.00\times$ & 100.00 & 85.60 & 93.45 & 95.55 & 93.65 \\
GHOST & 62.5\% & $2.67\times$ & 99.80 & 39.40 & 96.10 & 94.30 & 82.40 \\
GHOST & 75\% & $4.00\times$ & 99.40 & 22.60 & 89.95 & 94.85 & 76.70 \\
\bottomrule
\end{tabular*}
\end{table}

\begin{table}[H]
\centering
\footnotesize
\caption{Recall scores (\%) on four RULER retrieval tasks at 16K context for Nemotron Super. Traffic is the state-read plus state-write traffic reduction relative to Standard, using the same accounting as the accuracy tables. Average is the unweighted mean of the four task scores.}
\label{tab:recall-super}
\setlength{\tabcolsep}{3pt}
\begin{tabular*}{\textwidth}{@{\extracolsep{\fill}}llrrrrrr@{}}
\toprule
Method & Setting & Traffic & Single UUID & Multi-key & Multi-value & Multi-query & Average \\
\midrule
Standard & full-state & $1.00\times$ & 100.00 & 100.00 & 100.00 & 100.00 & 100.00 \\
ReplaySSM & full-state & $1.88\times$ & 100.00 & 100.00 & 100.00 & 100.00 & 100.00 \\
\midrule
SketchSSM & $\bar G=8$ & $9.39\times$ & 100.00 & 100.00 & 100.00 & 100.00 & 100.00 \\
SketchSSM & $\bar G=4$ & $11.84\times$ & 100.00 & 100.00 & 100.00 & 100.00 & 100.00 \\
\midrule
GHOST & 50\% & $2.00\times$ & 100.00 & 90.20 & 99.60 & 99.65 & 97.36 \\
GHOST & 62.5\% & $2.67\times$ & 99.40 & 44.80 & 93.65 & 95.20 & 83.26 \\
GHOST & 75\% & $4.00\times$ & 96.60 & 1.20 & 57.90 & 56.40 & 53.03 \\
\bottomrule
\end{tabular*}
\end{table}

\begin{table}[H]
\centering
\footnotesize
\caption{Recall scores (\%) on four RULER retrieval tasks at 16K context for Qwen3.8 Flash-Next. Traffic is the state-read plus state-write traffic reduction relative to Standard, using the same accounting as the accuracy tables. Average is the unweighted mean of the four task scores.}
\label{tab:recall-qwen}
\setlength{\tabcolsep}{3pt}
\begin{tabular*}{\textwidth}{@{\extracolsep{\fill}}llrrrrrr@{}}
\toprule
Method & Setting & Traffic & Single UUID & Multi-key & Multi-value & Multi-query & Average \\
\midrule
Standard & full-state & $1.00\times$ & 100.00 & 100.00 & 100.00 & 100.00 & 100.00 \\
ReplaySSM & full-state & $1.88\times$ & 100.00 & 100.00 & 100.00 & 100.00 & 100.00 \\
\midrule
SketchSSM & $\bar G=8$ & $10.68\times$ & 100.00 & 100.00 & 100.00 & 100.00 & 100.00 \\
SketchSSM & $\bar G=4$ & $12.81\times$ & 100.00 & 100.00 & 100.00 & 100.00 & 100.00 \\
\midrule
DRRQR & 50\% & $2.00\times$ & 99.80 & 100.00 & 99.75 & 100.00 & 99.89 \\
DRRQR & 62.5\% & $2.67\times$ & 99.40 & 98.60 & 97.10 & 99.25 & 98.59 \\
DRRQR & 75\% & $4.00\times$ & 65.20 & 21.60 & 33.90 & 51.50 & 43.05 \\
\bottomrule
\end{tabular*}
\end{table}

\begin{table}[H]
\centering
\footnotesize
\caption{Recall scores (\%) on four RULER retrieval tasks at 16K context for GLM 5.3 Flash. Traffic is the state-read plus state-write traffic reduction relative to Standard, using the same accounting as the accuracy tables. Average is the unweighted mean of the four task scores.}
\label{tab:recall-glm}
\setlength{\tabcolsep}{3pt}
\begin{tabular*}{\textwidth}{@{\extracolsep{\fill}}llrrrrrr@{}}
\toprule
Method & Setting & Traffic & Single UUID & Multi-key & Multi-value & Multi-query & Average \\
\midrule
Standard & full-state & $1.00\times$ & 99.80 & 100.00 & 97.80 & 100.00 & 99.40 \\
ReplaySSM & full-state & $1.88\times$ & 99.80 & 100.00 & 98.05 & 100.00 & 99.46 \\
\midrule
SketchSSM & $\bar G=8$ & $10.68\times$ & 100.00 & 100.00 & 98.95 & 100.00 & 99.74 \\
SketchSSM & $\bar G=4$ & $12.81\times$ & 100.00 & 100.00 & 98.80 & 100.00 & 99.70 \\
\midrule
GHOST & 50\% & $2.00\times$ & 100.00 & 98.80 & 82.10 & 100.00 & 95.23 \\
GHOST & 62.5\% & $2.67\times$ & 99.80 & 95.00 & 73.20 & 96.55 & 91.14 \\
GHOST & 75\% & $4.00\times$ & 61.20 & 47.20 & 47.30 & 31.40 & 46.78 \\
\bottomrule
\end{tabular*}
\end{table}

\FloatBarrier

\clearpage
\section{Ablation of Preserving Full-State Updates with Approximate Reads}
\label{app:read-only-comparison}

We evaluate the effect of preserving the full recurrent
state for updates while using a compact representation
only for reads. To this end, we construct ablation variants
within our proposed design that differ in their read
representation: a quantized state copy, a low-rank state
approximation, or a sketch.
The low-rank baseline builds on classical truncated
SVD~\citep{Eckart_Young_1936}, using a fixed basis
obtained offline from the leading eigenvectors of the
average state Gram matrix, with direct query projection
for coefficients. Unlike the sketch, this variant uses fixed rather than state-dependent coefficients.

As discussed in Section~\ref{sec:bg-state-access}, compressing
the recurrent state itself introduces errors that propagate
through subsequent state updates. Our design instead retains
the full state for these updates and applies approximation
only to the read path. Because updates occur only at flush
steps, their full-state traffic is amortized over multiple
decode steps.

Each variant constructs its compact read representation
from the updated state at a flush step and reads it between
flushes.
All variants retain the full state for subsequent updates;
they are ablations of the read representation within our design.

Table~\ref{tab:read-only-ablation} reports results for
quantized read-only copies, low-rank state representations, and sketches,
with state quantization included as a reference.
Retaining the full state for updates substantially improves
accuracy over quantizing the recurrent state itself.
Within this design, using a sketch further improves average
accuracy over reading a quantized state copy at comparable
state-access traffic reductions.
\begin{table}[H]
\centering
\scriptsize
\caption{Accuracy ablation for maintaining full states and approximating
reads on Nemotron Nano 9B v2 ($W=16$). Ours decouples full-state updates
from approximate reads and varies only the read representation.
Traffic is the state-read plus state-write
traffic reduction relative to Standard; sketch reads store the
sketch rows and coefficient map in BF16. Low-rank traffic includes
BF16 reads of both the compressed state and fixed projection map,
using the same per-head ranks as the sketch.}
\label{tab:read-only-ablation}
\setlength{\tabcolsep}{2pt}
\begin{tabular*}{\textwidth}{@{\extracolsep{\fill}}lllrrrrrr@{}}
\toprule
Method & Update & Read & Traffic & MATH-500 & AIME25 & GPQA-D & LCB & Average \\
\midrule
Baseline & full state & full state & $1.00\times$ & 97.33 & 71.67 & 59.43 & 68.99 & 74.36  \\
\midrule
Baseline & 6-bit state & 6-bit state & $5.06\times$ & 70.60 & 8.30 & 27.80 & 12.70 & 29.85  \\
Ours & full state & 4-bit state copy & $7.95\times$ & 97.40 & 68.33 & 55.05 & 63.49 & 71.07  \\
Ours & full state & low-rank state, $\bar G=8$ & $9.94\times$ & 94.93 & 57.92 & 49.83 & 45.61 & 62.07  \\
Ours (SketchSSM) & full state & sketch, $\bar G=8$ & $9.94\times$ & 97.07 & 71.67 & 60.77 & 68.15 & 74.41  \\
\midrule
Baseline & 4-bit state & 4-bit state & $7.40\times$ & 9.40 & 0.00 & 0.00 & 5.40 & 3.70  \\
Ours & full state & 2-bit state copy & $10.36\times$ & 88.20 & 30.00 & 39.39 & 26.98 & 46.14  \\
Ours & full state & low-rank state, $\bar G=4$ & $12.26\times$ & 88.13 & 34.58 & 45.96 & 22.65 & 47.83  \\
Ours (SketchSSM) & full state & sketch, $\bar G=4$ & $12.26\times$ & 96.73 & 70.42 & 58.25 & 62.33 & 71.93  \\
\bottomrule
\end{tabular*}
\end{table}

\FloatBarrier

\end{document}